\documentclass[10pt,twocolumn,letterpaper]{article}

\usepackage{cvpr}              

\usepackage[utf8]{inputenc} 
\usepackage[T1]{fontenc}    
\usepackage{hyperref}       
\usepackage{url}            
\usepackage{booktabs}       
\usepackage{amsfonts}       
\usepackage{nicefrac}       
\usepackage{microtype}      
\usepackage{xcolor}         

\usepackage{amsmath}
\usepackage{epsfig}
\usepackage{multirow}
\usepackage{amssymb}
\usepackage{pifont}
\usepackage{wrapfig}
\usepackage{comment}
\usepackage{subcaption}
\usepackage{placeins}
\usepackage{float}
\usepackage{tikz}
\usetikzlibrary{spy}
\usepackage{mathtools}
\usepackage{acro}

\DeclareAcronym{nf}{
  short = NF ,
  long = Neural Field
}

\DeclareAcronym{df}{
  short = DF ,
  long = Distance Field
}

\DeclareAcronym{ndf}{
  short = NDF ,
  long = Neural Distance Field
}

\DeclareAcronym{sdf}{
  short = SDF ,
  long = Signed Distance Field
}

\DeclareAcronym{tsdf}{
  short = TSDF ,
  long = Truncated Signed Distance Field
}

\DeclareAcronym{udf}{
  short = UDF ,
  long = Unsigned Distance Field
}

\DeclareAcronym{nn}{
  short = NN ,
  long = Neural Network
}

\DeclareAcronym{rdae}{
  short = RDAE ,
  long = Rate-Distortion Autoencoder
}

\DeclareAcronym{dpm}{
  short = DPM ,
  long = Diffusion Probabilistic Model
}

\DeclareAcronym{cd}{
  short = CD ,
  long = Chamfer Distance
}

\DeclareAcronym{emd}{
  short = EMD ,
  long = Earth Mover Distance
}

\DeclareAcronym{pc}{
  short = PC ,
  long = Point Cloud
}

\DeclareAcronym{mc}{
  short = MC ,
  long = Marching Cubes
}

\DeclareAcronym{vpcc}{
  short = VPCC ,
  long = Video-based Point Cloud Compression
}

\DeclareAcronym{gpcc}{
  short = GPCC ,
  long = Geometry-based Point Cloud Compression
}

\DeclareAcronym{mlp}{
  short = MLP ,
  long = Multilayer Perceptron
}

\DeclareAcronym{psnr}{
  short = PSNR ,
  long = Peak Signal-to-Noise-Ratio
}

\DeclareAcronym{pcgcv2}{
  short = PCGCv2 ,
  long = Point Cloud Geometry Compression v2
}

\DeclareAcronym{vqad}{
  short = VQAD ,
  long = Vector Quantized Autodecoder
}

\newcommand{\stanford}{Stanford shape repository}
\newcommand{\garments}{MGN dataset}
\newcommand{\attributes}{8i Voxelized Full Bodies}
\newcommand{\attr}{8iVFB}

\newcommand{\fourPlotsSize}{0.24\textwidth}
\newcommand{\fourPlotsSizeWithLabel}{0.255\textwidth}
\newcommand{\stanfordPCQualitativeFigWidth}{0.09\textwidth}
\newcommand{\attributesPCQualitativeFigWidth}{0.1\textwidth}

\definecolor{cvprblue}{rgb}{0.21,0.49,0.74}

\def\paperID{9534} 
\def\confName{CVPR}
\def\confYear{2024}

\title{3D Compression Using Neural Fields}

\newcommand{\authorDist}{4mm}
\author{Janis Postels$^1$ \hspace{\authorDist} Yannick Strümpler$^2$ \hspace{\authorDist} Klara Reichard$^3$ \hspace{\authorDist} Luc Van Gool$^1$ \hspace{\authorDist} Federico Tombari$^{2,3}$\\
$^1$ETH Zurich \hspace{\authorDist} $^2$Google \hspace{\authorDist} $^3$Technical University Munich\\
\tt\small \{jpostels, vangool\}@vision.ee.ethz.ch \hspace{4mm} klara.reichard@tum.de \\ \tt\small\{strumpler, tombari\}@google.com
}

\begin{document}
\maketitle

\begin{abstract}
Neural Fields (NFs) have gained momentum as a tool for compressing various data modalities - \eg images and videos. This work leverages previous advances and proposes a novel NF-based compression algorithm for 3D data. We derive two versions of our approach - one tailored to watertight shapes based on Signed Distance Fields (SDFs) and, more generally, one for arbitrary non-watertight shapes using Unsigned Distance Fields (UDFs). We demonstrate that our method excels at geometry compression on 3D point clouds as well as meshes. Moreover, we show that, due to the NF formulation, it is straightforward to extend our compression algorithm to compress both geometry and attribute (\eg color) of 3D data.         
\end{abstract}  

\section{Introduction}\label{sec:intro}

It becomes increasingly important to transmit 3D content over band-limited channels. Consequently, there is a growing interest in algorithms for compressing related data modalities~\cite{schwarz2018emerging, quach2022survey}. 
Compared to image and video compression, efforts to reduce the bandwidth footprint of 3D data modalities have gained less attention. 
%
%
Moreover, the nature of 3D data renders it a challenging problem. 
Typically, image or video data live on a well-defined regular grid. However, the structure, or \textit{geometry}, of common 3D data representations such as \acp{pc} and meshes only exists on a lower-dimensional manifold embedded in the 3D world.
%
Moreover, this is often accompanied by \textit{attributes} that are only defined on the geometry itself.   

Notably, the MPEG group has recently renewed its call for standards for 3D compression identifying point clouds as the central modality~\cite{mpeg2017call,schwarz2018emerging}. To this end, geometry and attribute compression are identified as the central constituents. 
\ac{gpcc} and \acp{vpcc} have emerged as standards for compressing 3D \acp{pc} including attributes~\cite{schwarz2018emerging}.
\ac{gpcc} is based on octrees and Region-Adaptive Hierarchical Transforms (RAHT)~\cite{de2016compression} and \ac{vpcc} maps geometry and attributes onto a 2D regular grid and applies state-of-the-art video compression algorithms. 
%
%
Subsequently, there has been a growing effort in developing methods for compressing either the geometry, attributes or both simultaneously~\cite{cao20193d, quach2022survey}. 

%
\acp{nf} have recently been popularized for a variety of data modalities including images, videos and 3D~\cite{10.1111:cgf.14505}. To this end, a signal is viewed as a scalar- or vector-valued field on a coordinate space and parameterized by a neural network, typically a \ac{mlp}. Interestingly, there is a growing trend of applying \acp{nf} to compress various data modalities, \eg images~\cite{dupont2021coin, strumpler2022implicit}, videos~\cite{chen2021nerv, zhangimplicit, lee2022ffnerv} or medical data~\cite{dupont2022coin++}. Hereby, the common modus operandi is to overfit an \ac{mlp} to represent a signal, \eg image/video, and, subsequently, compress its parameters using a combination of quantization and entropy coding. Our work 
proposes the first \ac{nf}-based 3D compression algorithm. 
%
%
In contrast to other geometry compression methods~\cite{schwarz2018emerging,quach2022survey}, \acp{nf} have been demonstrated to represent 3D data regardless of whether it is available in form of \acp{pc}~\cite{atzmon2020sal,atzmon2020sald} or meshes~\cite{park2019deepsdf,chibane2020neural}.
%
%
\acp{nf} do not explicitly encode 3D data, but rather implicitly in form of \acp{sdf}~\cite{park2019deepsdf}, \acp{udf}~\cite{chibane2020neural} or vector fields~\cite{rella2022neural}.
%
%
Therefore, one typically applies marching cubes~\cite{lorensen1987marching,guillard2022meshudf} on top of distances and signs/normals obtained from the \ac{nf} to extract the geometry. 

%
We show that \ac{nf}-based compression using \acp{sdf} leads to state-of-the-art geometry compression.
%
%
As \acp{sdf} assume watertight shapes, a general compression algorithm requires \acp{udf}. However, vanilla \acp{udf} lead to inferior compression performance since the non-differentiable target requires increased model capacity. To mitigate this, we apply two impactful modifications to \acp{udf}. 
Specifically, we apply a suitable activation function to the output of \acp{udf}. Further, we regularize \acp{udf} trained on \acp{pc}.
Therefore, we tune the distribution from which training points are sampled 
and apply an $\ell1$-penalty on the parameters.
Lastly, we demonstrate that \acp{nf} are not only a promising approach for compressing the geometry of 3D data but also its attributes by viewing attributes as a vector-valued field on the geometry.

\section{Related Work}\label{sec:related_work}
\subsection{Modelling 3D Data Using Neural Fields}

%
\acp{nf} were initially introduced to 3D shape modelling in the form of occupancy fields~\cite{mescheder2019occupancy,chen2019learning} and \acp{sdf}~\cite{park2019deepsdf}. 
3D meshes are extracted from the resulting \acp{sdf} using marching cubes~\cite{lorensen1987marching}. Further, Chibane \etal~\cite{chibane2020neural} proposed Neural Unsigned Distance fields to represent 3D shapes using \acp{udf} and, thus, allow for modeling non-watertight shapes. They obtain shapes as \acp{pc} by projecting uniformly sampled points along the negative gradient direction of the resulting \acp{udf}. Later, Guillard \etal\ introduced MeshUDF~\cite{guillard2022meshudf} building on \ac{mc}~\cite{lorensen1987marching} which denotes a differentiable algorithm for converting \acp{udf} into meshes. We instantiate our method with both \acp{sdf} and \acp{udf} leading to a more specialized and a, respectively, more general version. More recently, Rella \etal\ proposed to parameterize the gradient field of \acp{udf} with a neural network. Regarding the architecture of the \ac{mlp} used for parameterizing \acp{nf}, Tancik \etal~\cite{tancik2020fourier} solidify that positional encodings improve the ability of coordinate-based neural networks to learn high frequency content. Further, Sitzmann \etal~\cite{sitzmann2020implicit} demonstrate that sinusoidal activation functions have a similar effect. In this work we utilize sinusoidal activation functions as well as positional encodings. 

\subsection{Compression Using Neural Fields}

Recently, there has been an increasing interest in compressing data using \acp{nf} due to promising results and their general applicability to any coordinate-based data. Dupont \etal~\cite{dupont2021coin} were the first to propose \acp{nf} for image compression. Subsequently, there was a plethora of work extending this seminal work. Strümpler \etal~\cite{strumpler2022implicit} improved image compression performance and encoding runtime by combining SIREN~\cite{sitzmann2020implicit} with positional encodings~\cite{tancik2020fourier} and applying meta-learned initializations~\cite{tancik2021learned}. Schwarz \etal~\cite{schwarzmeta} and Dupont \etal~\cite{dupont2022coin++} further expand the idea of meta-learned initializations for \ac{nf}-based compression. Furthermore, various more recent works have improved upon \ac{nf}-based image compression performance~\cite{ladune2022cool, gao2022sinco, damodaran2023rqat}. Besides images, \ac{nf}-based compression has been extensively applied to videos~\cite{chen2021nerv, zhangimplicit, rho2022neural, kimscalable, maiya2022nirvana, lee2022ffnerv}. Despite the recent interest in \ac{nf}-based compression, its application to compressing 3D data modalities remains scarce. Notably,
there has been an increasing effort to compress 3D scenes by compressing the parameters of Neural Radiance Fields (NeRF)~\cite{bird20213d, isik2021neural, takikawa2022variable, li2023compress}. However, this work directly compresses 3D data modalities (\acp{pc}/meshes) while NeRF-compression starts from 2D image observations.

\subsection{3D Data Compression}

Typically, 3D compression is divided into geometry and attribute compression. We refer to Quach \etal~\cite{quach2022survey} for a comprehensive survey. 
%

\textbf{Geometry Compression.} MPEG has identified \acp{pc} - including attributes - as a key modality for transmitting 3D information~\cite{schwarz2018emerging}. Subsequently, it introduced \ac{gpcc} and \ac{vpcc} for compressing the geometry and attributes captured in 3D \acp{pc}. \ac{gpcc} is a 3D native algorithm which represents \acp{pc} using an efficient octree representation for geometry compression. On the other hand, \ac{vpcc} maps the geometry and attributes onto a 2D grid and, then, takes advantage of video compression algorithms. Moreover, Draco~\cite{galligan2018google} allows compressing \acp{pc} and meshes. For mesh compression it relies on the edge-breaker algorithm~\cite{rossignac1999edgebreaker}. Tang \etal~\cite{tang2018real} take a different approach by extracting the \ac{sdf} from a 3D geometry and then compressing it.

Early works on learned geometry compression use a \ac{rdae} based on 3D convolutions~\cite{guarda2019deep, quach2019learning, guarda2020deep, quach2020improved}. 
Wang \etal~\cite{wang2021lossy} also apply 3D convolutions to \ac{pc} compression and later introduce an improved multi-scale version based on sparse convolutions, \ie{} \ac{pcgcv2}~\cite{wang2021multiscale}.
\ac{pcgcv2} improves upon \ac{pc} geometry compression using 3D convolutions in prior works. Thus, we use it as a learned baseline in \autoref{ssec:geometry_compression}.
SparsePCGC~\cite{wang2022sparse} further improves upon \ac{pcgcv2}~\footnote{Note that it is not possible to compare directly to SparsePCGC due to the absence of training code/model checkpoints.}.
%
%
Tang \etal~\cite{tang2020deep} compress watertight shapes including color attributes using \acp{tsdf}. Hereby, the signs of the \ac{tsdf} are compressed losslessly using a learned conditional entropy model, the \ac{udf} is encoded/decoded using 3D convolutions and texture maps are compressed using a custom tracking-free UV parameterization. In contrast to all prior work on learned 3D compression, we overfit a single \ac{mlp} to parameterize a single signal. While this increases the encoding time, it also drastically renders our method less vulnerable to domain shifts. Moreover, in contrast to Tang \etal{} which focuses on \acp{sdf}, we also utilize \acp{udf} to compress non-watertight geometries. 
NGLOD~\cite{takikawa2021neural} proposed to represent 3D data using feature grids with variable resolution in a parameter efficient manner. VQAD~\cite{takikawa2022variable} further substantially improves upon NGLOD by quantizing these feature grids.


\textbf{Attribute Compression.} \ac{gpcc}~\cite{schwarz2018emerging} compresses attributes using Region-Adaptive Hierarchical Transforms (RAHT)~\cite{de2016compression}, while \ac{vpcc} maps attributes onto 2D images and applies video compression algorithms. Further, Quach \etal~\cite{quach2020folding} propose a folding-based \acp{nn} for attribute compression. Tang \etal~\cite{tang2020deep} introduce a block-based UV parameterization and, then, applies video compression similar to \ac{vpcc}. Isik \etal~\cite{isik2021lvac} demonstrate that vector-valued \acp{nf} are a promising tool for attribute compression. 
%
In contrast, this work tackles both geometry and attribute compression.
Sheng \etal~\cite{sheng2021deep} and Wang \etal~\cite{wang2022sparse} compress point cloud attributes using a \ac{rdae} based on PointNet++~\cite{qi2017pointnet++} and, respectively, 3D convolutions. 

\section{Method}\label{sec:method}
\begin{figure*}[t!]
    \centering
    \includegraphics[width=0.8\textwidth]{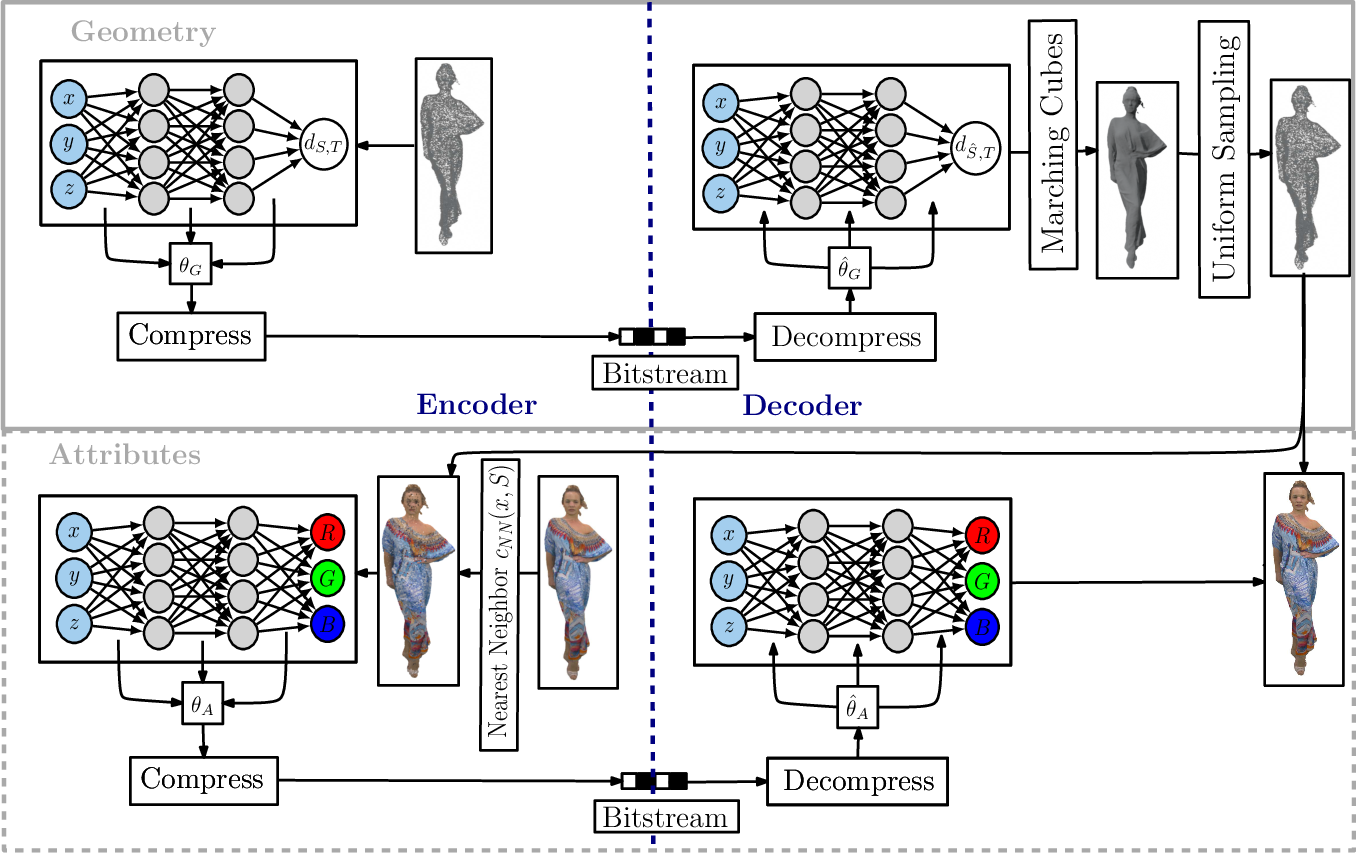}
    \caption{Overview of \ac{nf}-based 3D compression. Geometry compression is in the upper part (see \autoref{ssec:geom_tndf}), while the lower part visualizes attribute compression (optional) (see \autoref{ssec:attr_tnf}). Encoding is shown on the left side. It contains two parts: a) geometry representation using \acp{nf} (see \autoref{ssec:geom_tndf}) and, resp., attribute representation (see \autoref{ssec:attr_tnf}) and b) the \ac{nf} parameter compression (see \autoref{ssec:compressing_nfs}).}
    \label{fig:method_vis}
\end{figure*}

%
Generally, we fit a single \ac{nf} to a 3D shape comprised of a \ac{pc}/mesh and, optionally, another \ac{nf} to the attributes (\eg color) on the geometry. 
Then, we compress the parameters of the \ac{nf}. 
Specifically, \autoref{ssec:geom_tndf} describes how we model the geometry of 3D data - for \acp{pc} as well as meshes - using truncated \acp{ndf}. 
Further, we explain how meshes and, ultimately, \acp{pc} can be recovered from \acp{df}. 
\autoref{ssec:attr_tnf} elaborates on additionally compressing attributes (\eg color) of 3D data. 
Lastly, \autoref{ssec:compressing_nfs} describes our approach to compressing the parameters of \acp{nf} representing the underlying 3D data.
\autoref{fig:method_vis} outlines our geometry and attribute compression pipeline. 

\subsection{Representing Geometries with Truncated Neural Distance Fields}\label{ssec:geom_tndf}

We represent 3D geometries implicitly using \acp{df}. 
%
%
A \ac{df} is a scalar field $\textrm{DF}: \mathbb{R}^3 \mapsto \mathbb{R}$ that for a given 3D geometry assigns every point $x\in\mathbb{R}^3$ the distance $d_{S}(x)\in\mathbb{R}_{\geq 0}$ to the closest point $x_{S}\in\mathbb{R}^3$ on the surface $S$ of the geometry. 
We refer to such scalar fields as \acp{udf} and omit the dependence on $x$, $d_S \coloneqq d_S(x)$.
If the underlying geometry is watertight, we can further define the signed distance $d_{S}\in\mathbb{R}$ which is negative inside $S$ and positive on the outside. 
These instances are termed \acp{sdf}.
In both cases, the surface $S$ is implicitly defined by the level set $\{ x \vert d_{S} = 0 \}$ of the \ac{df}. 
In \autoref{sec:experiments} we demonstrate that using \acp{udf} leads to strong compression performance while being generally applicable. 
However, when handling watertight shapes, \acp{sdf} yield further significant improvements.

\textbf{Truncated Neural Distance Fields.} We parameterize $d_S$ using \acp{nn} $NF_{\theta_G}$ with parameters $\theta_G$ - in particular \acp{mlp} mapping coordinates to the corresponding values of the scalar field similar to recent work on \acp{nf}~\cite{park2019deepsdf, sitzmann2020implicit, chibane2020neural}. 
Our goal is to learn compressed representations of 3D geometries in form of the parameters $\theta_G$ and, thus, it is important to limit the number of parameters. 
To this end, we do not train \acp{ndf} to parameterize the entire \ac{df} $d_S$ but rather a truncated version of it. 
Hence, we intuitively only store the information in the \ac{df} that is necessary to recover the 3D geometry. 
Such a  truncated \ac{df} $d_{S,T}$ is characterized by a maximal distance $d^* > 0$ and defined as 
\begin{equation*}
  d_{S,T} = 
    \begin{cases}
      d_{S} & \textrm{ if } \vert d_{S,T} \vert \leq  d^*\\
      \textrm{sgn}(d_{S})d \in \{ \delta \in \mathbb{R}: \vert \delta \vert > d^*\} & \textrm{ else.}
   \end{cases}
\end{equation*}
where $\textrm{sgn}(d_{S})$ returns the sign of $d_{S}$. We only require $\vert d_{S,T} \vert $ to be larger than $d^*$ but do not fix its value. 
Thus, the \acp{ndf} can represent the 3D geometry with fewer parameters by focusing the model's capacity on the region closest to the surface.

\textbf{Architecture.} We use sinusoidal activation functions in the \acp{mlp}~\cite{sitzmann2020implicit} combined with positional encodings~\cite{tancik2020fourier}.
This has been shown to improve the robustness of \acp{nf} to quantization~\cite{strumpler2022implicit}. 
Chibane \etal~\cite{chibane2020neural} originally proposed to parameterize \acp{udf} using \acp{mlp} with a ReLU activation function for the output of the last layer to enforce $NF_{\theta_G} \geq 0$ $\forall x\in \mathbb{R}^3$. 
In contrast, we apply $\textrm{abs}(x) = \left| x \right|$ which drastically improves performance in the regime of small models (see \autoref{ssec:geometry_compression}). 
This originates from the fact that, unlike ReLU, $\textrm{abs}(x)$ allows to correctly represent a \ac{udf} using negative values prior to the last activation function. 
This again increases the flexibility of the model. 
When modeling \acp{sdf}, we apply the identity as the final activation function.
More details are in the supplement.

\textbf{Optimization.} We train on a dataset of point-distance pairs. 
Following prior work~\cite{sitzmann2020implicit,chibane2020neural}, we sample points from a mixture of three distributions - uniformly in the unit cube, uniformly from the level set and uniformly from the level set with additive Gaussian noise of standard deviation $\sigma$. This encourages learning accurate distances close to the surface.
When training on \acp{pc}, we restrict ourselves to approximately uniformly distributed \acp{pc}.
Non-uniform \acp{pc} can be sub/super-sampled accordingly.
%
%
In contrast to prior work on compressing other data modalities using \acp{nf}~\cite{dupont2021coin,strumpler2022implicit,dupont2022coin++,chen2021nerv,lee2022ffnerv}, implicitly representing 3D geometries - in particular in the form of \acp{pc} - is susceptible to overfitting as extracting a mesh from \acp{df} using \ac{mc} requires the level set to form a 2D manifold.
A \ac{nf} trained on a limited number of points may collapse to a solution where the level set rather resembles a mixture of delta peaks.
We counteract overfitting using two methods.
We find that $\sigma$ is an important parameter for the tradeoff between reconstruction quality and generalization (see \autoref{ssec:geometry_compression}).
We find that for the distribution of natural shapes, the values $\sigma=0.01$ (\acp{sdf}) and $\sigma=0.025$ (\acp{udf}) work well across datasets.
Secondly, we penalize the $\ell 1$-norm of $\theta$. 
This further has the benefit of sparsifying the parameters $\theta$~\cite{tibshirani1996regression} and, consequently, rendering them more compressible~\cite{strumpler2022implicit}. 
Overall, we train \acp{nf} to predict the above truncated \acp{udf}/\acp{sdf} using the following loss function (with $\hat{d} = NF_{\theta_G}(x)$):

\begin{equation}\label{eq:total_loss_geometry}
    \mathcal{L}_{G}(\theta_G) = \mathcal{L}_{D}(\theta_G) + \lambda_{\ell 1}\left|\left|\theta_G\right|\right|_1
\end{equation}

where $\mathcal{L}_{D}(\theta_G) = \textrm{E}\left[ \left(\hat{d} - \textrm{sgn}(d_{S})\min(\vert d_{S} \vert,d^*)\right)^2 \right] $ if $\vert d_{S} \vert \leq d^* \text{ or } \vert \hat{d} \vert \leq d^*$ and 0 otherwise.

%
%

\textbf{Extracting Geometries from Distance Fields.} 
Our compressed representation implicitly encodes the 3D surface. For comparison and visualization purposes, we need to convert it to an explicit representation, namely a \ac{pc} or mesh as part of our decoding step.
Obtaining a uniformly sampled \ac{pc} directly from a \ac{df} is non-trivial.
Hence, we initially convert the \acp{df} into meshes in both \ac{pc} compression and mesh compression scenarios. 
In the case of \acp{sdf}, we apply \ac{mc}~\cite{lorensen1987marching} to obtain a mesh of the 3D geometry.
Further, we extract meshes from \acp{udf} using the recently proposed differentiable \ac{mc} variant MeshUDF~\cite{guillard2022meshudf}.
Note that Chibane \etal~\cite{chibane2020neural} originally extracted points by projecting uniform samples along the gradient direction of \acp{udf}.
However, this leads to undesirable clustering and holes on geometries containing varying curvature.
When compressing \acp{pc}, we further sample points uniformly from the extracted meshes.
Notably, this is the primary reason for the inability of our compression algorithm to perform lossless compression of \acp{pc} - even in the limit of very large \acp{mlp}.
However, it achieves state-of-the-art performance in the regime of strong compression ratios across various datasets on 3D compression - using \acp{pc}/meshes with/without attributes (see \autoref{sec:experiments}).
%
%
Further, sampling \acp{pc} from the shape's surface fundamentally limits the reconstruction quality in terms of \ac{cd}. 
However, unlike previous methods that approximately memorize the original \ac{pc} directly, our method learns the underlying geometry.

\subsection{Representing 3D Attributes with Neural Fields}\label{ssec:attr_tnf}

Besides the geometry of 3D data, we further compress its attributes (\eg color) using \acp{nf}. 
To this end, we follow the high level approach of other attribute compression methods and compress the attributes given the geometry~\cite{quach2022survey}.
Thus, after training an \ac{mlp} to represent the geometry of a particular 3D shape we train a separate \ac{nf} with parameters $\theta_A$ to correctly predict attributes on the approximated surface $\hat{S}$ of the geometry $x\in\{x|NF_{\theta_G}(x)=0\}$.
Therefore, for a given point $x$ on $\hat{S}$ we minimize the $\ell 2$-distance to the attribute $c_{NN}(x, S)$ of the nearest neighbour on the true surface $S$:
\begin{equation*}\label{eq:total_loss_attributes}
    \mathcal{L}_{A}(\theta_A) = \underset{x\in \hat{S}}{\textrm{E}}\left[ \left(NF_{\theta_A}(x) - c_{NN}(x, S) \right)^2 \right] + \lambda_{\ell 1}\left|\left|\theta_A\right|\right|_1
\end{equation*}
$\lambda_{\ell 1}$ represents the strength of the regularization of $\theta_A$ and $c_{NN}(x, S) = c( \operatorname*{argmin}_{x' \in S} \Vert x - x'\Vert_2 ) $ with $c(\cdot)$ extracting the attribute at a surface point. Alternatively, one may also optimize a single \ac{nf} to jointly represent a geometry and its attributes. 
However, then the \ac{nf} has to represent attributes in regions $x\notin \hat{S}$ which wastes capacity.
The supplement contains an empirical verification.

\subsection{Compressing Neural Fields}\label{ssec:compressing_nfs}

In the proposed compression algorithm $\theta_G$, and optionally $\theta_A$, represent the 3D data. 
Therefore, it is important to further compress these using \ac{nn} compression techniques. 
We achieve this by first quantizing $\theta_{G/A}$, retraining the quantized \ac{mlp} to regain the lost performance and, lastly, entropy coding the resulting quantized values.
Subsequently, we describe each step in detail.

\textbf{Quantization.} We perform scalar quantization of $\theta_{G/A}$ using a global bitwidth $b$ which corresponds to $2^b$ possible values. 
We use a separate uniformly-spaced quantization grid for each layer of the \ac{mlp}.
The layer-wise step size $s_l$ is defined by the $\ell_{\infty}$-norm of the parameters $\theta_{G/A}^l$ of layer $l$ and $b$
\begin{equation*}
    s_l = \frac{\left|\left|\theta_{G/A}^l\right|\right|_{\infty}}{2^b-1}
\end{equation*}
and has to be stored to recover the quantized values.
Note, that the quantization grid is centered around 0 - where $\theta_{G/A}$ peaks - to improve the gain of lossless compression using entropy coding.

\textbf{Quantization-Aware Retraining.} We perform a few epochs of quantization-aware retraining with a much smaller learning rate.
We compute gradients during quantization-aware training using the straight-through estimator~\cite{bengio2013estimating}.
We also experimented with solely training \acp{nf} using quantization-aware optimization.
However, this drastically decreased convergence speed and, thus, increased the encoding time without improving performance.

\textbf{Entropy Coding.} Finally, we further losslessly compress the quantized parameters $\hat{\theta}_{G/A}^l$ using a near optimal general purpose entropy coding algorithm\footnote{\url{https://github.com/google/brotli}}.


\section{Experiments}\label{sec:experiments}
\begin{figure*}[t!]
\centering
\setlength{\tabcolsep}{1pt}
\begin{tabular}{cccc}
\multicolumn{4}{c}{\includegraphics[width=1.0\textwidth]{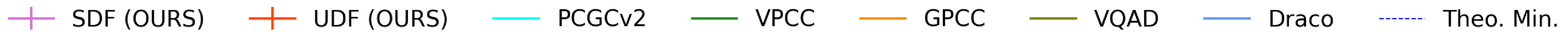}} \\
\includegraphics[width=\fourPlotsSizeWithLabel]{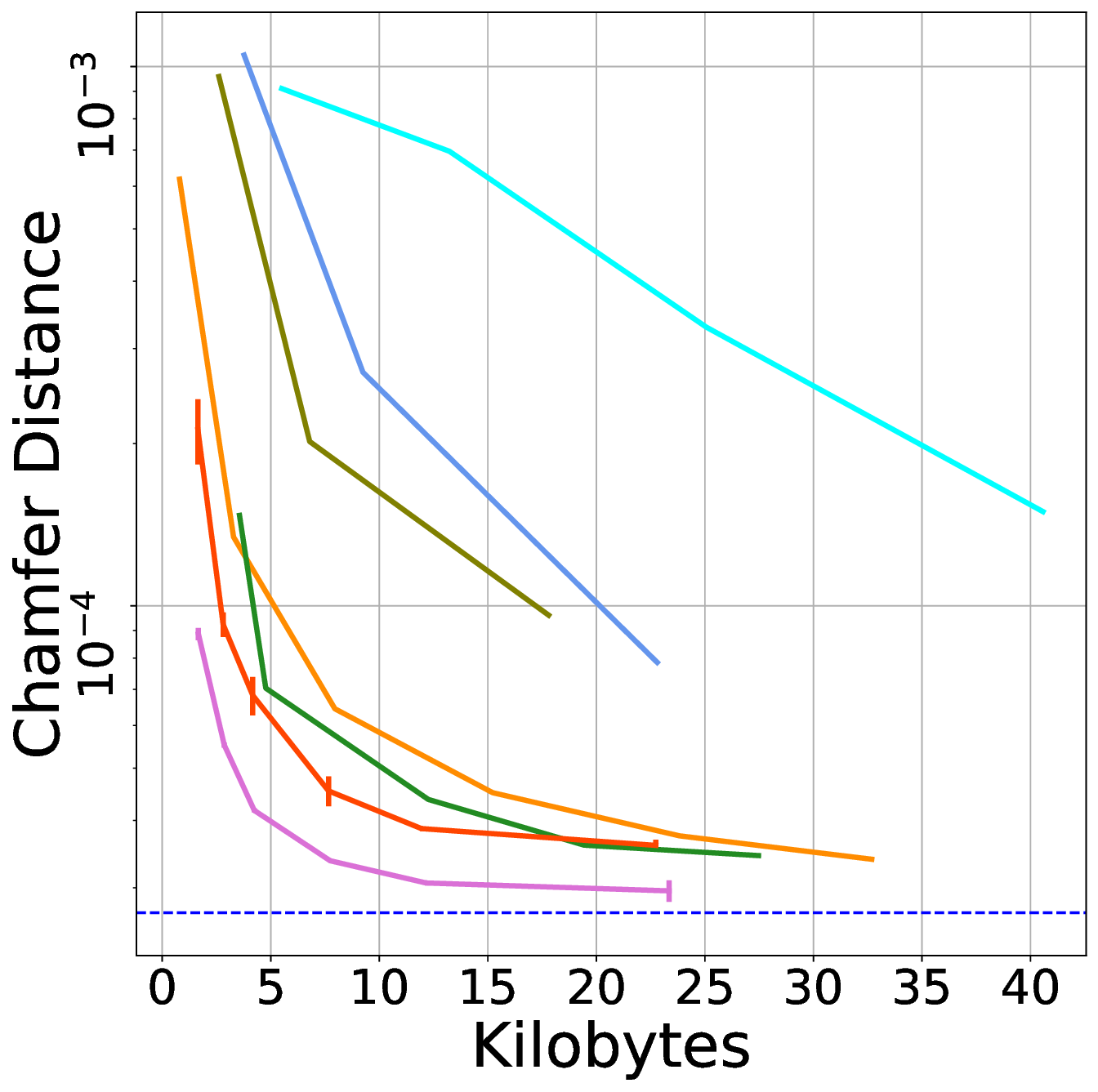} & 
\includegraphics[width=\fourPlotsSize]{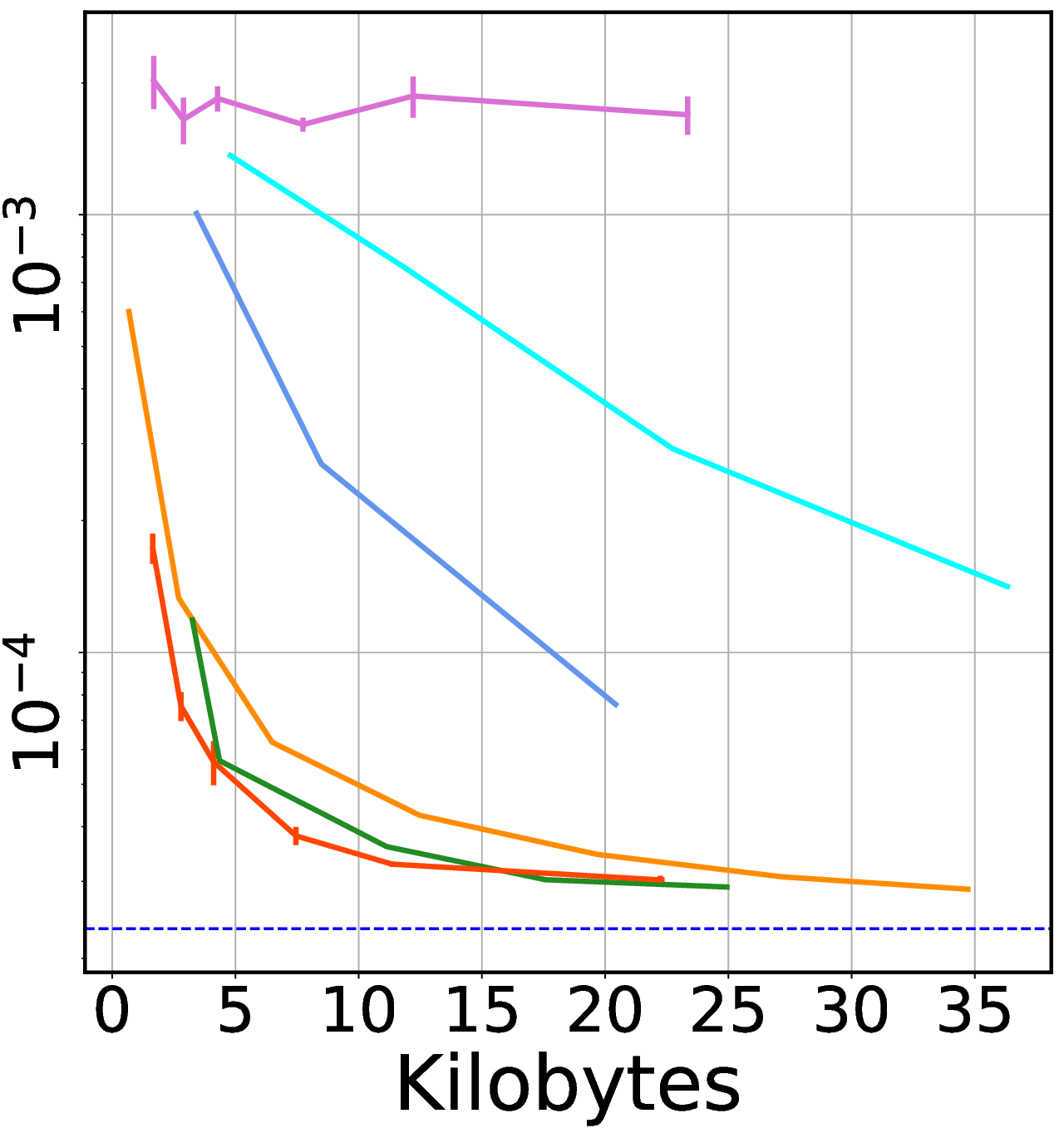} &
\includegraphics[width=\fourPlotsSize]{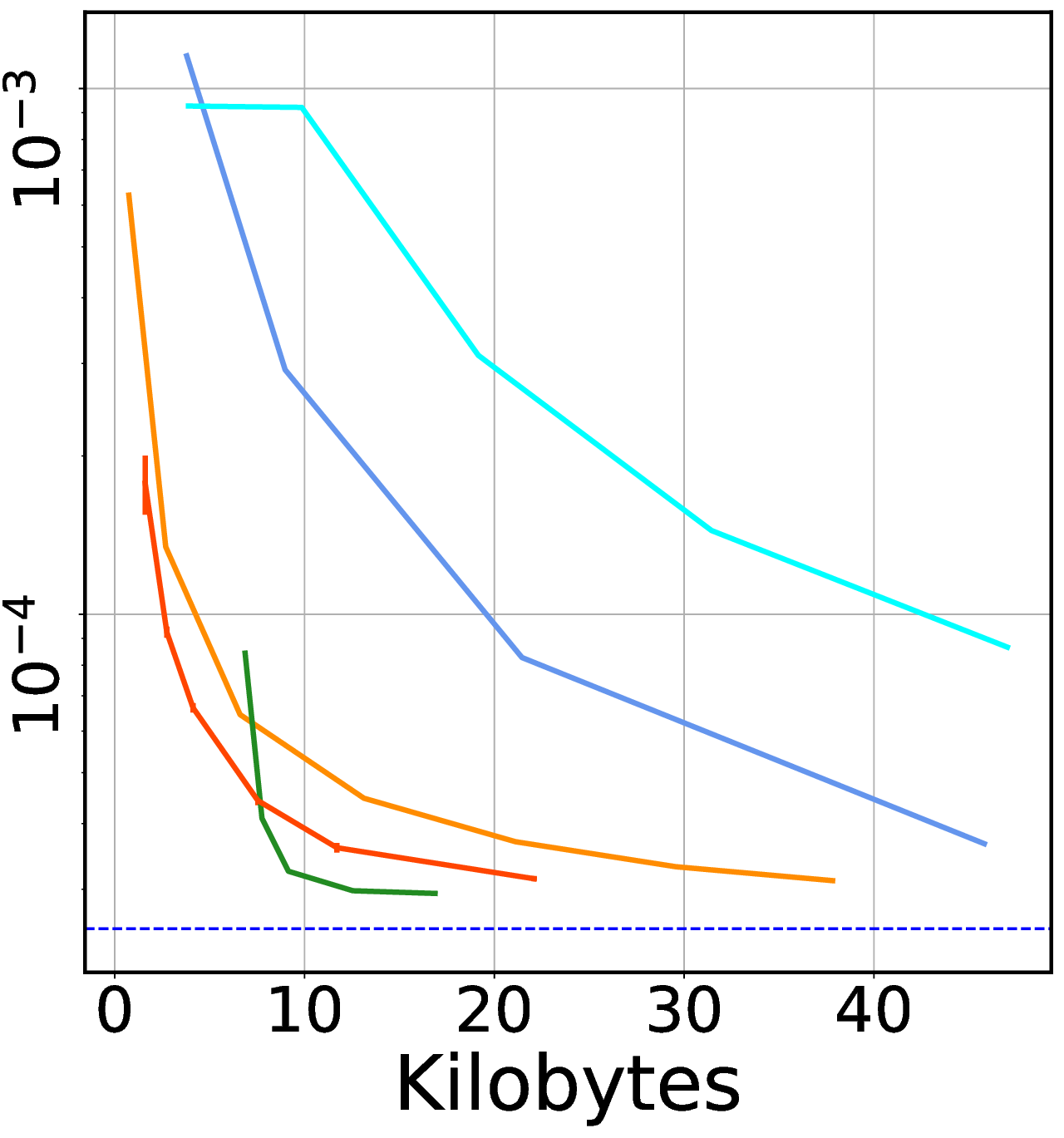} &
\includegraphics[width=\fourPlotsSize]{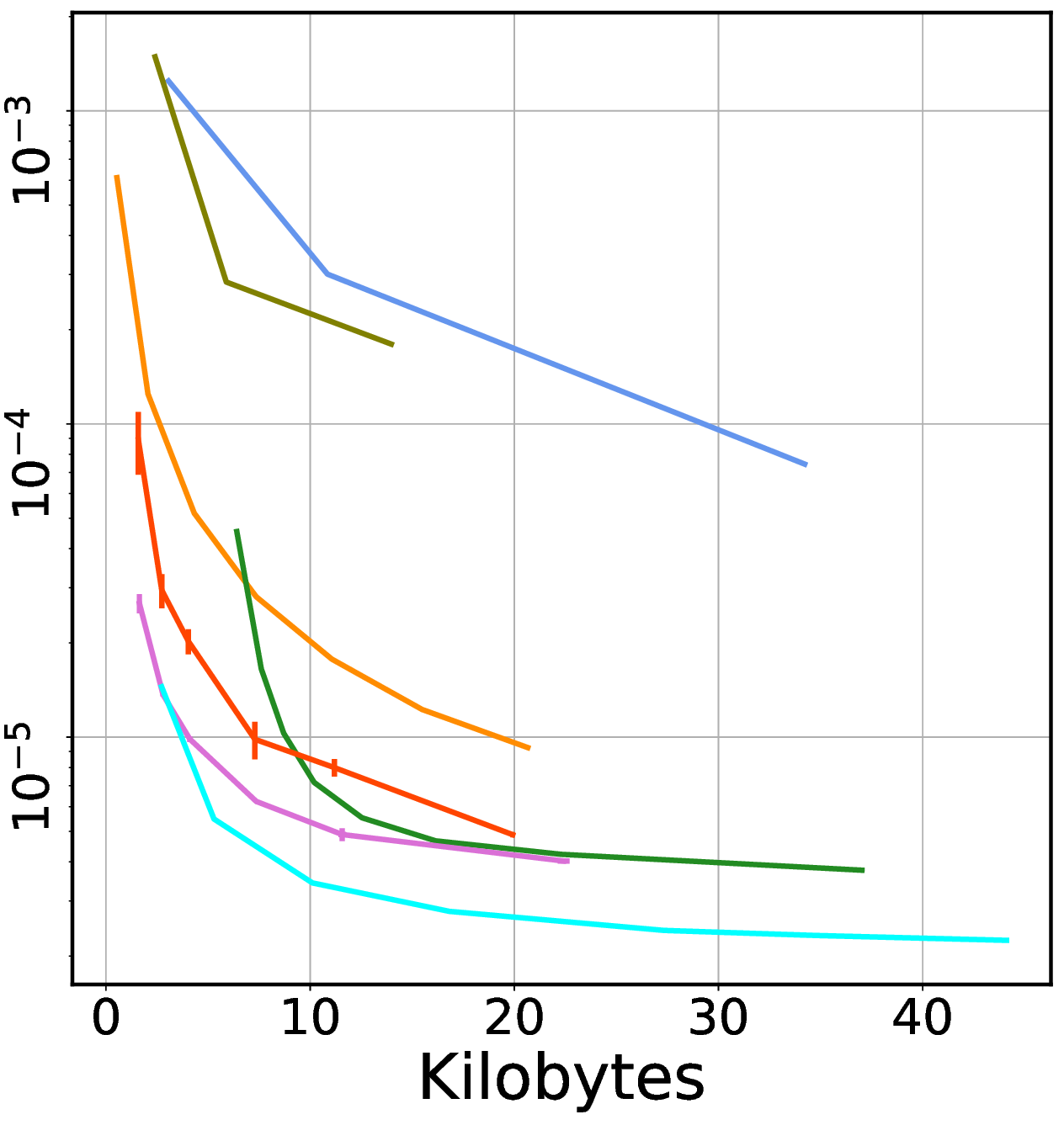} \\
a) Watertight shapes &  b) All shapes  & c) \garments{} &  d) \attr{} \\
\end{tabular}
\caption{Rate-distortion plot for \textbf{\ac{pc} compression} on the \stanford{}~\cite{turk1994zippered} (a/b), on the \garments{} ~\cite{bhatnagar2019mgn} (c) and \attr{}~\cite{d20178i} (d) depicting the average \ac{cd} and number kilobytes for \acp{nf} based on \acp{udf}/\acp{sdf}, \ac{pcgcv2}, VQAD, Draco, GPCC and VPCC. \ac{pcgcv2}/VQAD are learned neural baselines, and Draco, \ac{gpcc} and \ac{vpcc} are non-learned standards. On the \stanford{} we report performance on a subset only containing watertight shapes (a) and and all shapes (b). We do not evaluate the performance of VQAD and \acp{nf} using \acp{sdf} on \garments{} since it is exclusively comprised of non-watertight shapes. There is not theo. min. in (d) as we operate directly on \acp{pc}.}\label{fig:geometry_pc_stanford_garments_attributes}
\end{figure*}

%
\autoref{ssec:geometry_compression} depicts experiments on geometry compression - both for \acp{pc} and meshes. 
Moreover, \autoref{ssec:geometry_compression} analyses the impact of components our compression algorithm. 
\autoref{ssec:attribute_compression} investigates the performance on 3D geometry and attributes compression. 
We exclusively consider color attributes. 

\textbf{Datasets.} We conduct experiments on three datasets. 
Firstly, we evaluate geometry compression - \ac{pc} as well as mesh compression - on a set of shapes from the \stanford{}~\cite{turk1994zippered} and a subset of the \garments{}~\cite{bhatnagar2019mgn} which was also used by Chibane \etal~\cite{chibane2020neural}. 
The former are high quality meshes consisting of both watertight and non-watertight shapes. 
The latter are lower quality meshes of clothing. 
Moreover, we conduct experiments on \acp{pc} extracted from \attributes{} (\attr)~\cite{d20178i}. 
\attr{} consists of four sequences of high quality colored \acp{pc}. Each \ac{pc} contains between 700,000 and 1,000,000 points. 
We use the first frame of each sequence in our experiments. 
We refer to the supplement for visualizations of each dataset.

\textbf{Data Preprocessing.} We center each 3D shape around the origin. 
Then, we scale it by dividing by the magnitude of the point (\ac{pc}), resp. vertex (mesh), with largest distance to the origin. 
For \acp{pc}, we compute the ground truth distance $d_S$ using the nearest neighbour in the \ac{pc}. 
For meshes, we use a CUDA implementation to convert them to \acp{sdf}~\cite{hao2020dualsdf} which we further adapt to generate \acp{udf}. 
We train on points from a mixture of three distributions. 
20\% are sampled uniformly $x\sim \left[ -1, 1 \right]^3$, 40\% are sampled uniformly from the surface $S$ and the remaining 40\% are sampled uniformly from $S$ and perturbed by additive Gaussian noise $\mathcal{N}(0;\sigma)$. 
For \acp{sdf} we set $\sigma=0.01$ and for \acp{udf} $\sigma=0.025$ if not stated otherwise. 
We sample 100,000 points from $S$ in our experiments on geometry compression on the \stanford{} and the \garments{} and we use all points in the ground truth \ac{pc} on \attr{}.
%
%
Color attributes are translated and scaled to the interval $\left[ -1, 1 \right]$.

\textbf{Evaluation and Metrics.} We evaluate the reconstruction quality of geometry compression using the \ac{cd}. 
The \ac{cd} is calculated between the ground truth \ac{pc} and the reconstructed \ac{pc} when handling \acp{pc}.
For mesh compression, we report the    \ac{cd} between \acp{pc} uniformly sampled from the ground truth and reconstructed mesh.
If not stated otherwise, we use 100,000 points on the \stanford{} and the \garments{}, and all available points on \attr{}. 
We evaluate the quality of reconstructed attributes using a metric based on the \ac{psnr}.
Therefore, we compute the \ac{psnr} between the attribute of each point in the ground truth \ac{pc} and its nearest neighbour in the reconstructed \ac{pc}, and vice versa.
The final metric is then the average between both \acp{psnr}.
Subsequently, we simply refer to this metric as \ac{psnr}.
Following Strümpler \etal{}~\cite{strumpler2022implicit}, we traverse the Rate-Distortion curve by varying the width $\in \{ 16, 24, 32, 48, 64, 96 \}$ of the \ac{mlp}. 

\textbf{Baselines.} We compare \ac{nf}-based 3D compression with the non-learned baselines Draco~\cite{galligan2018google}, \ac{gpcc}~\cite{schwarz2018emerging} and \ac{vpcc}~\cite{schwarz2018emerging}. 
\ac{vpcc}, which is based on video compression, is the non-learned state-of-the-art on compressing 3D \acp{pc} including attributes.
We compare our method with the learned neural baseline \ac{pcgcv2}~\cite{wang2021multiscale} which is the state-of-the-art \ac{rdae} based on 3D convolutions and VQAD~\cite{takikawa2022variable} which builds quantized hierarchical feature grids. 
None of the baselines supports all data modalities/tasks used in our experiments. 
Geometry compression on meshes is only supported by Draco. 
On geometry compression using \acp{pc}, we compare with all baselines. 
Lastly, joint geometry and attribute compression is only supported by \ac{gpcc} and \ac{vpcc} which we evaluate on \attr{}. Note that Draco supports normal but not color attribute compression.
When sampling from meshes, we also report the theoretical minimum, \ie the expected distance between independently sampled point sets.

\textbf{Optimization.} We train all \acp{nf} using a batch size of 10,000 for 500 epochs using a learning rate of $10^{-4}$ and the Adam optimizer~\cite{kingma2014adam}. 
We use $\lambda_{\ell 1}=10^{-8}$ and $d^*=0.1$. 
Each \ac{mlp} contains 2 hidden layers.
We follow Sitzmann \etal{}~\cite{sitzmann2020implicit} and use the factor 30 as initialization scale.
We use 16 fourier features as positional encodings on geometry compression and 8 on attribute compression.
\acp{nf} are quantized using a bitwidth $b=8$ and quantization-aware retraining is performed for 50 epochs using a learning rate of $10^{-7}$.
Each \ac{nf} is trained on a single V100 or A100.

\subsection{Geometry Compression}\label{ssec:geometry_compression}

We investigate \ac{nf}-based 3D geometry compression and compare it with the baselines. 
We evaluate our method on \ac{pc} and mesh compression and verify design choices.   

\begin{figure*}
\setlength{\tabcolsep}{6pt} 
\renewcommand{\arraystretch}{1} 
\centering
    \begin{tabular}{cccccccc}\includegraphics[width=\stanfordPCQualitativeFigWidth]{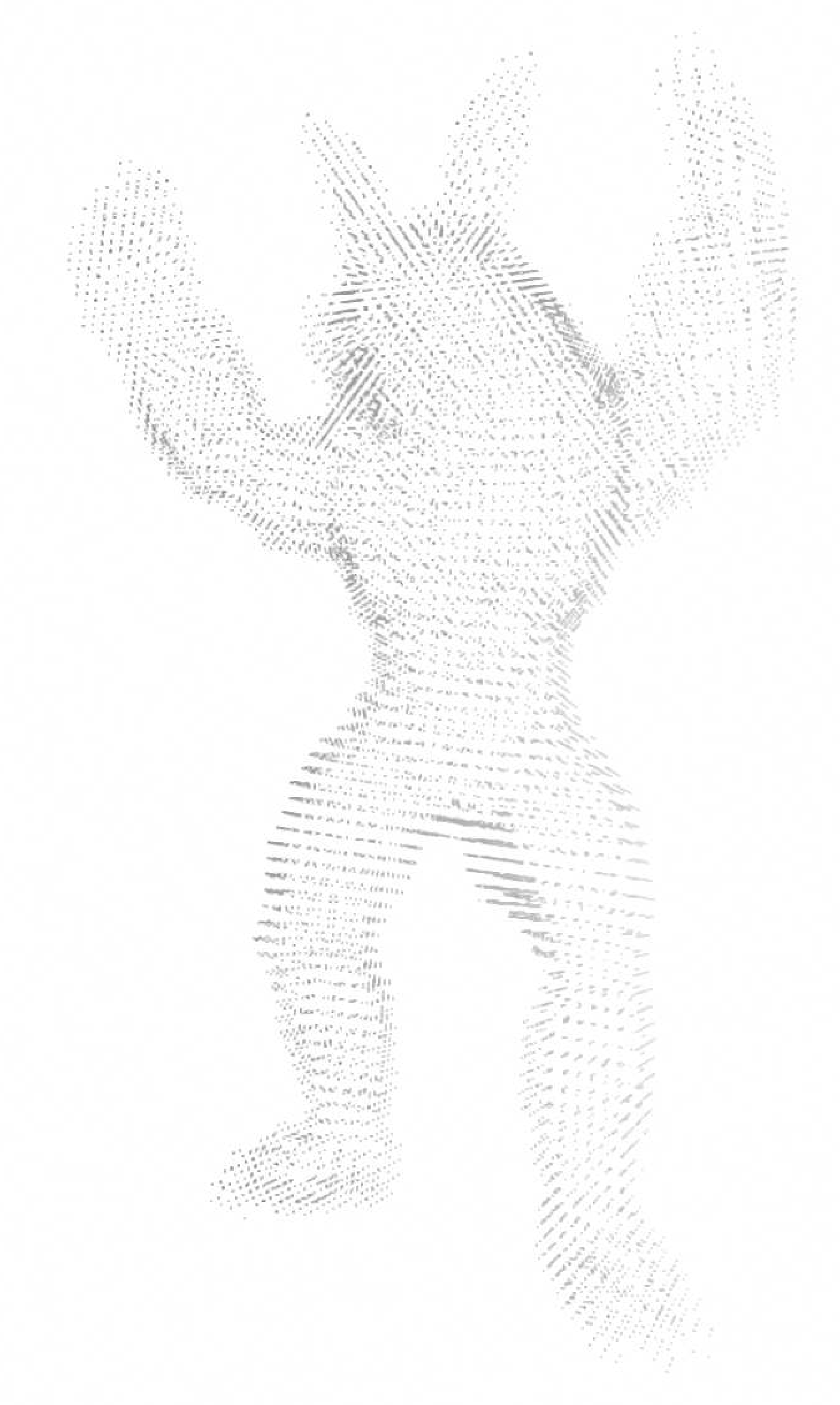} & 
    \includegraphics[width=\stanfordPCQualitativeFigWidth]{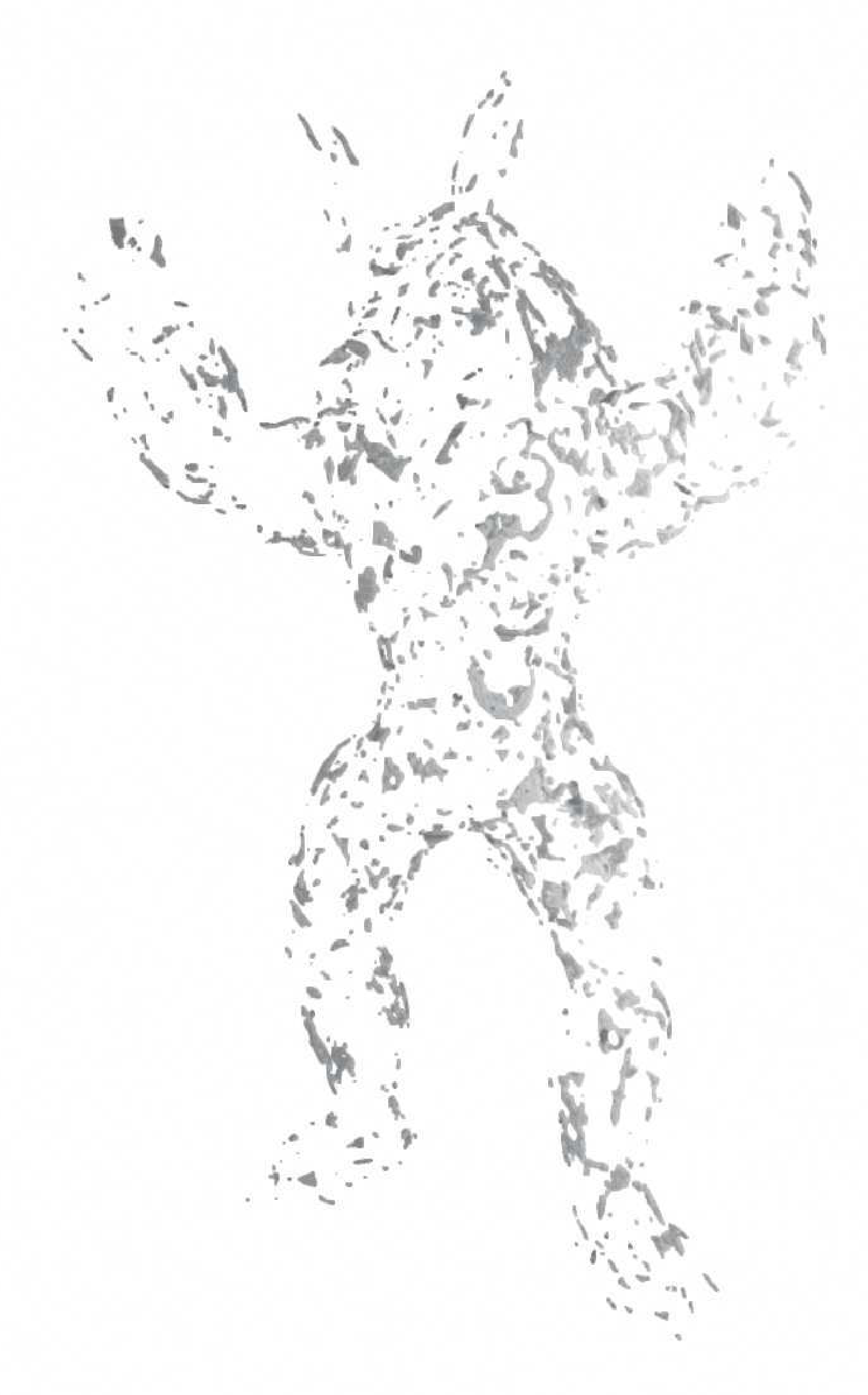} &
    \includegraphics[width=\stanfordPCQualitativeFigWidth]{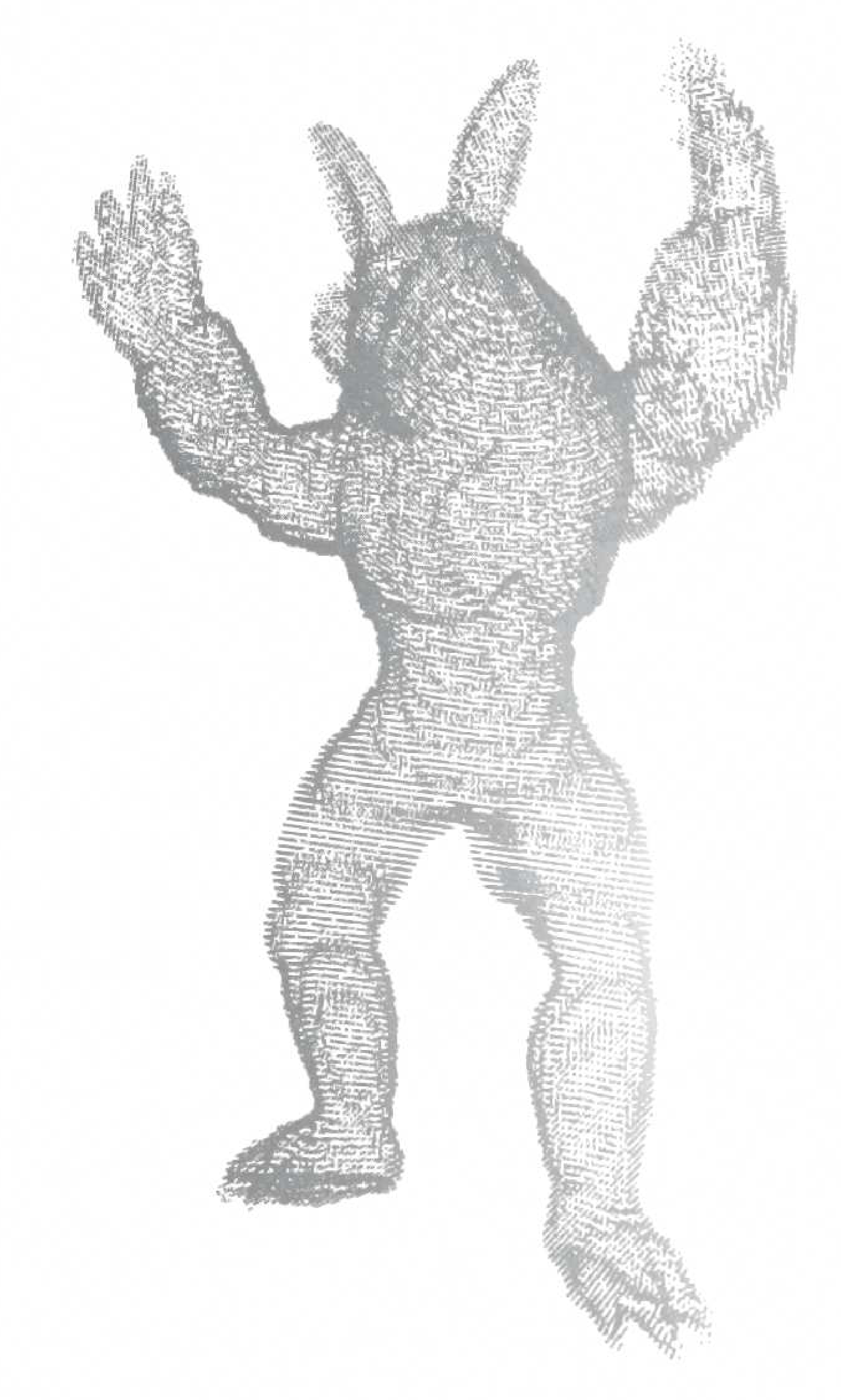} & 
    \includegraphics[width=\stanfordPCQualitativeFigWidth]{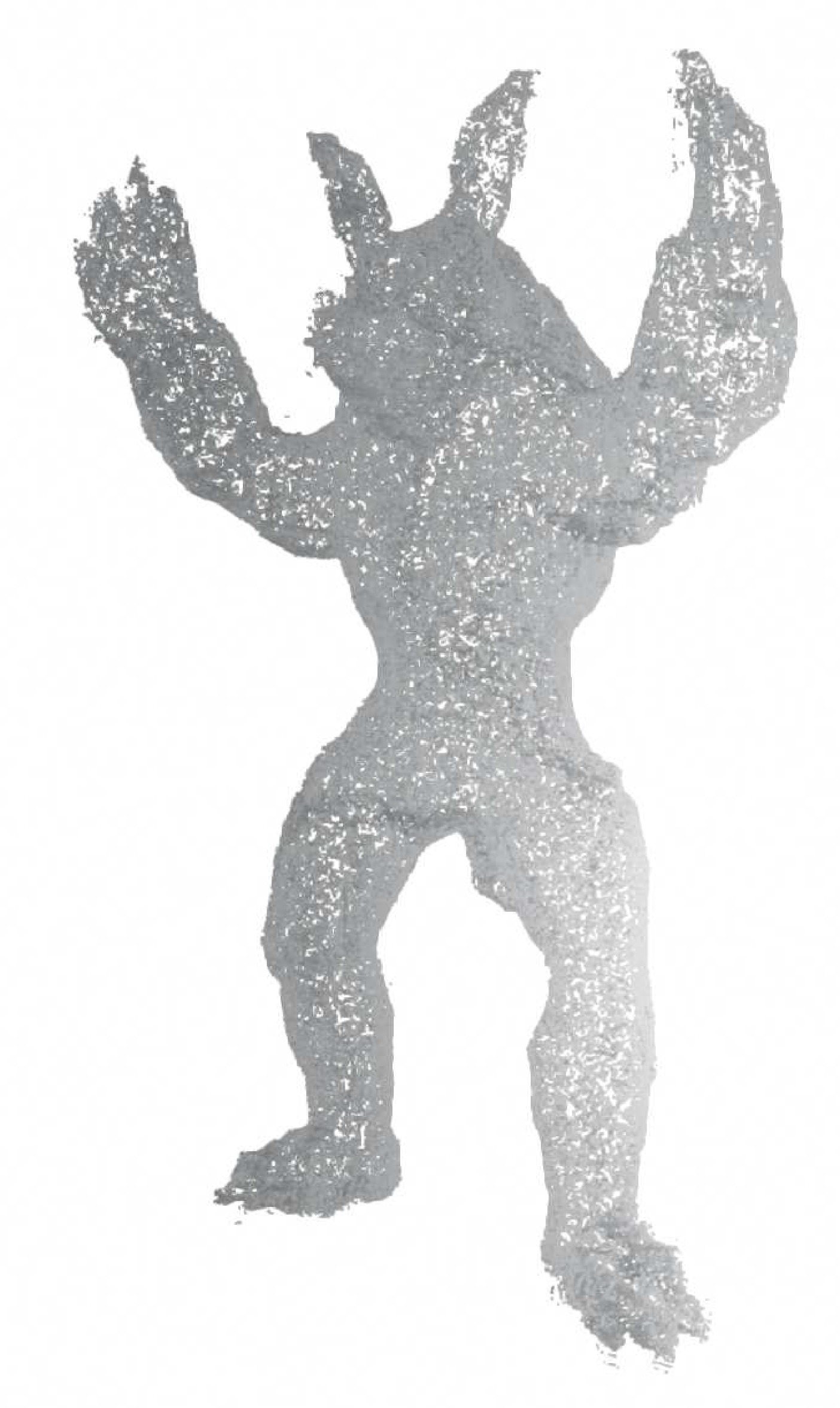} &
    \includegraphics[width=\stanfordPCQualitativeFigWidth]{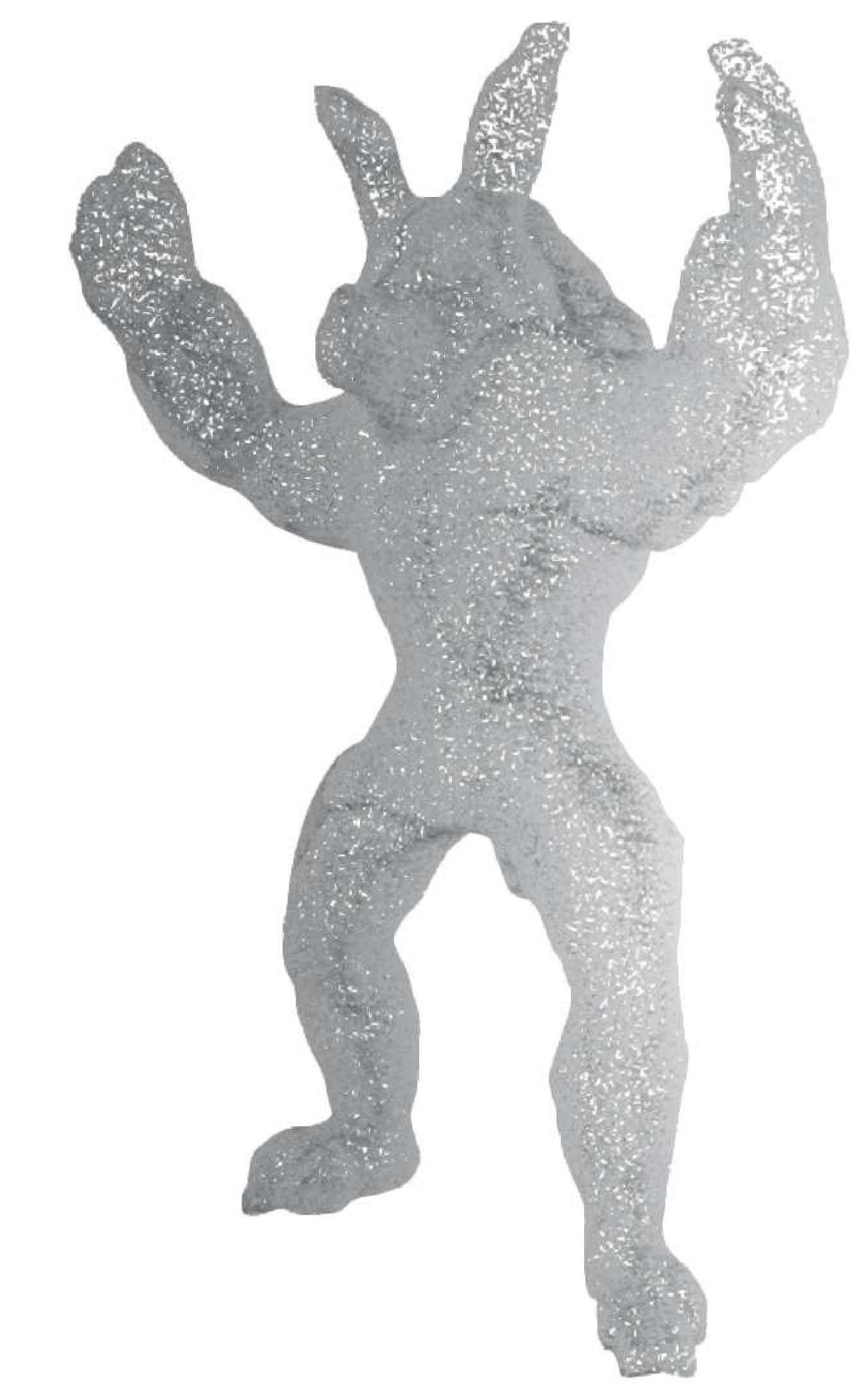} &
    \includegraphics[width=\stanfordPCQualitativeFigWidth]{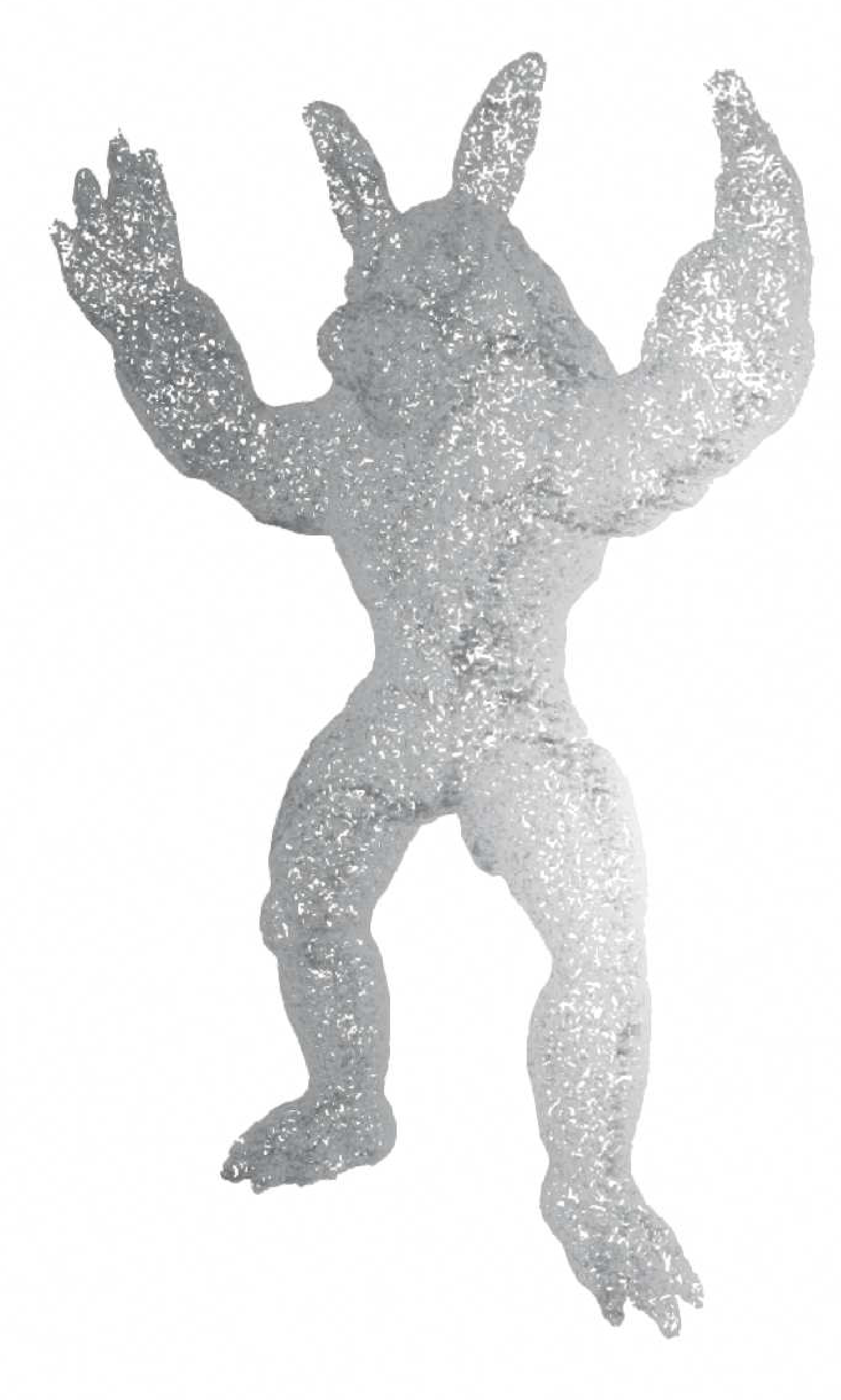} & 
    \includegraphics[width=\stanfordPCQualitativeFigWidth]{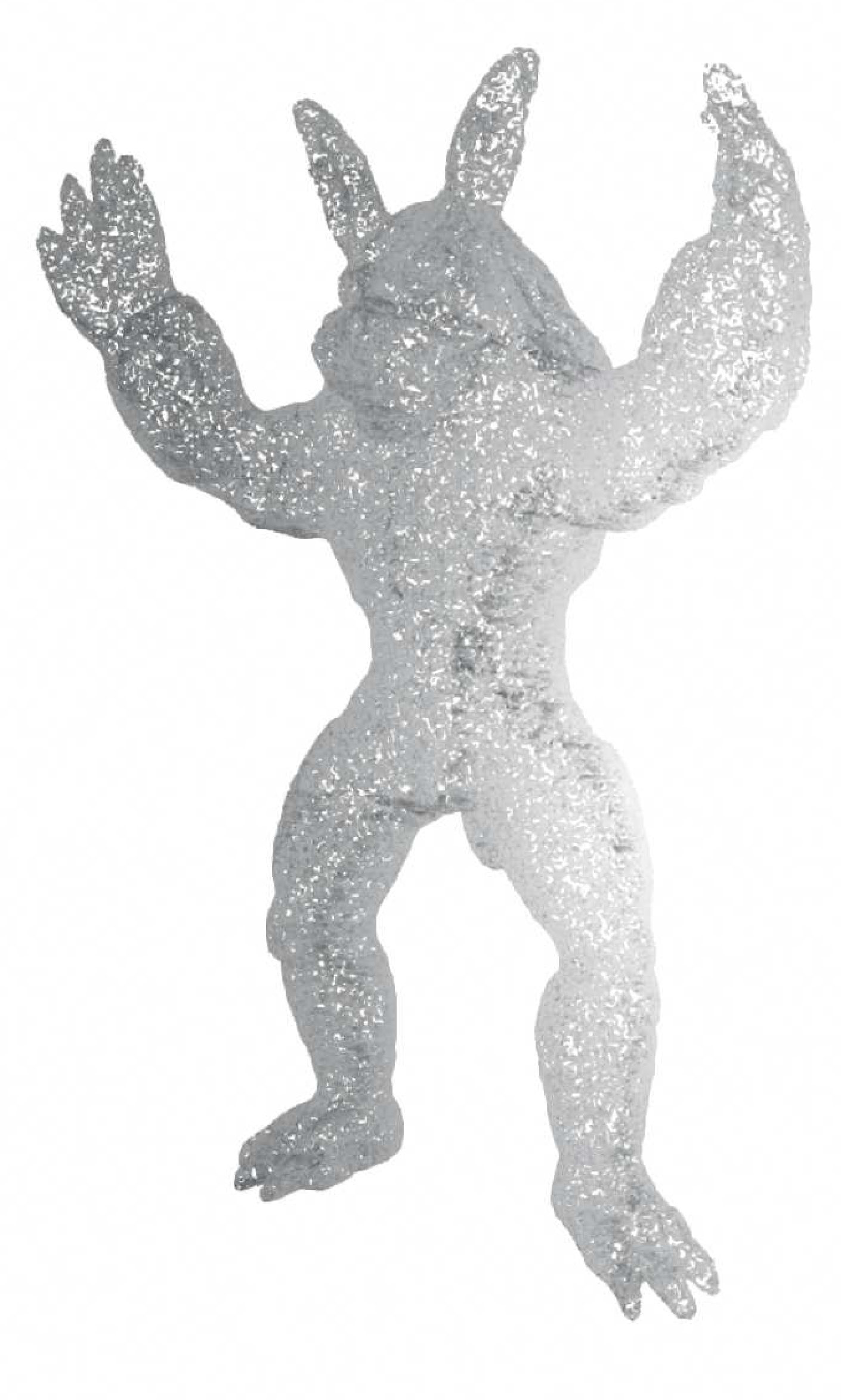} &
    \includegraphics[width=\stanfordPCQualitativeFigWidth]{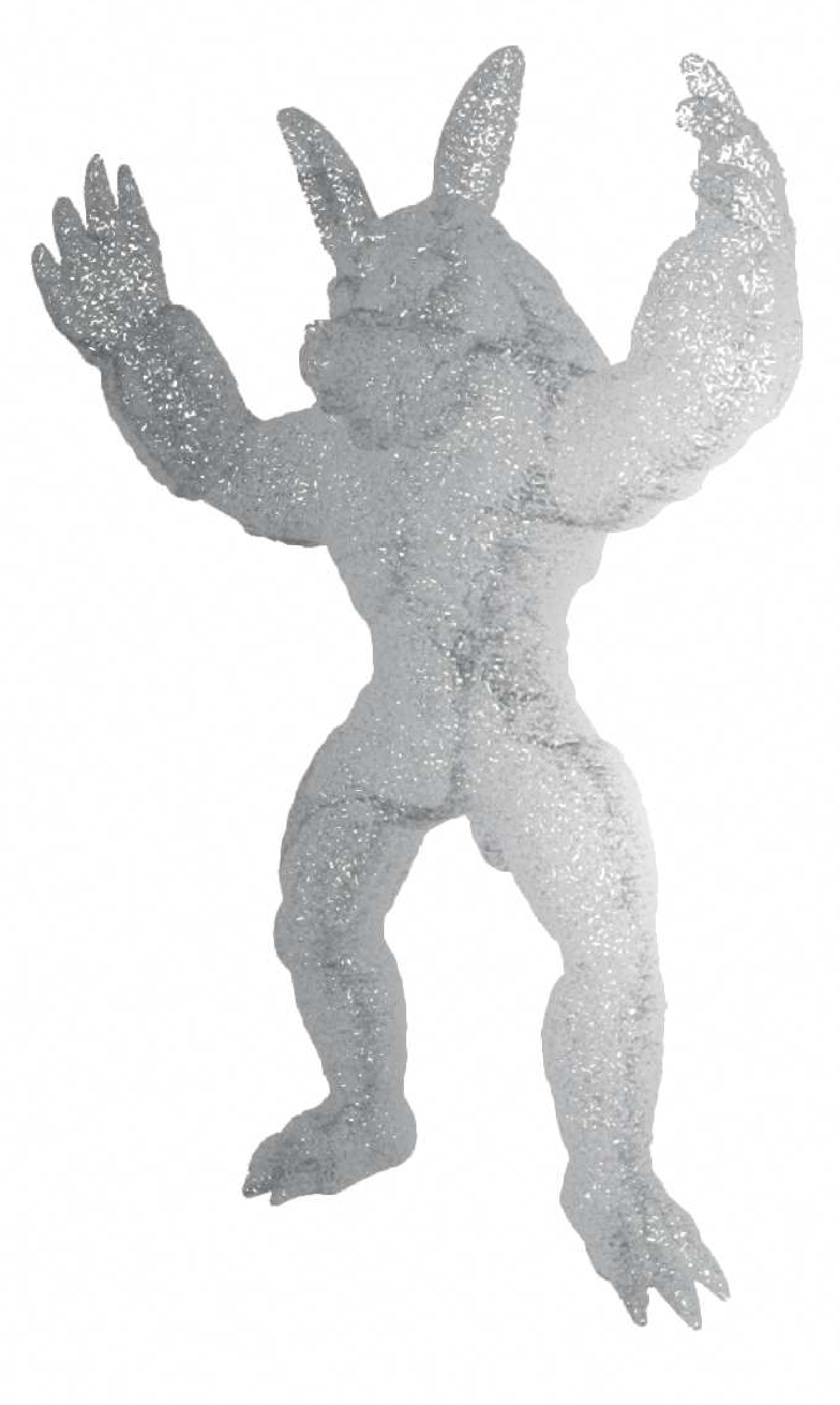}\\
    \small a) \small Draco & \small b) \ac{pcgcv2} & c) \small GPCC & \small d) VPCC & \small e) VQAD & \small f) UDF & \small g) SDF & \small h) GT\\
    \small 9.2 KB & \small 6.6 KB & \small 4.8 KB & \small 8.0 KB & \small 6.9 KB & \small 4.2 KB & \small 4.2 KB & \small 2.9 MB \\
    \end{tabular}
\caption{We depict qualitative results on "Armadillo" of the \stanford{} for Draco (a), \ac{pcgcv2} (b), \ac{gpcc} (c), \ac{vpcc} (d), \ac{vqad} (e), \acp{udf}/\acp{sdf} (f/g) and the ground truth (h).}\label{fig:stanford_geometry_qualitative}
\end{figure*}

\textbf{Point Clouds.} We evaluate \ac{pc} compression on the \stanford{}, the \garments{} and \attr{}. 
\autoref{fig:geometry_pc_stanford_garments_attributes} (a) \& (b) depict the result on the \stanford{}, where we show results on the watertight subset (a) and all shapes (b), and \autoref{fig:geometry_pc_stanford_garments_attributes} (c) \& (d) contain the results on the \garments{} and \attr{}.
Further, \autoref{fig:stanford_geometry_qualitative} shows qualitative results of reconstructed \acp{pc} on the \stanford{}.
We observe that for watertight shapes on the \stanford{}, \acp{sdf} outperform the baselines for all levels of compression.
On \attr{}, \ac{nf}-based compression is only outperformed by \ac{pcgcv2} whose performance drops steeply on other datasets.
\ac{pcgcv2} was trained on ShapeNet~\cite{shapenet2015} which contains artificial shapes with less details than the real scans in the \stanford{} and the \garments{}. 
Despite its strong performance on \attr{}, we hypothesize that \ac{pcgcv2} reacts very sensitive to the shift in the distribution of high frequency contents in the geometry.
%
%
%
As expected the performance of \acp{sdf} deteriorates when adding non-watertight shapes since the \ac{sdf} is not well defined in this case.
Further, \acp{udf} also outperform the baselines stronger compression ratios. 
%
Similarly, on the \garments{}, where we do not evaluate \acp{sdf} as all shapes are non-watertight, and on \attr{} \acp{udf} perform well for strong compression ratios and are only out performed by \ac{vpcc} on weaker compression ratios and by \ac{pcgcv2} on \attr{}. 


\begin{figure*}
\centering
\setlength{\tabcolsep}{1pt}
\begin{tabular}{cccc}
\multicolumn{4}{c}{\includegraphics[width=1.0\textwidth]{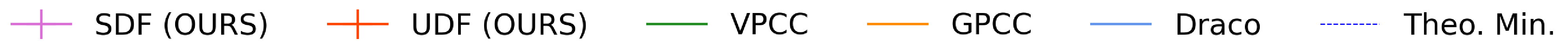}} \\
      \includegraphics[width=\fourPlotsSizeWithLabel]{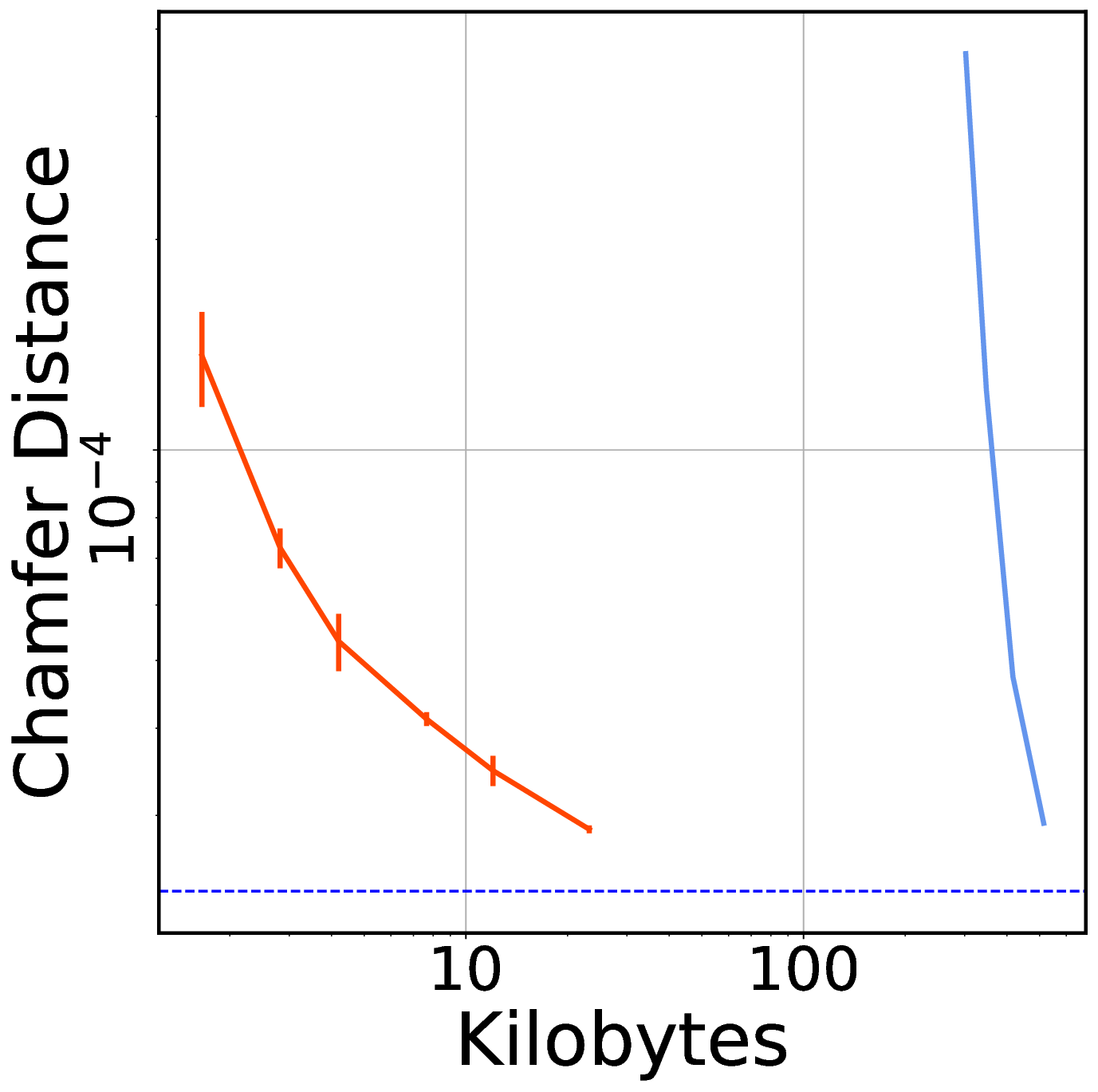} &
      \includegraphics[width=\fourPlotsSize]{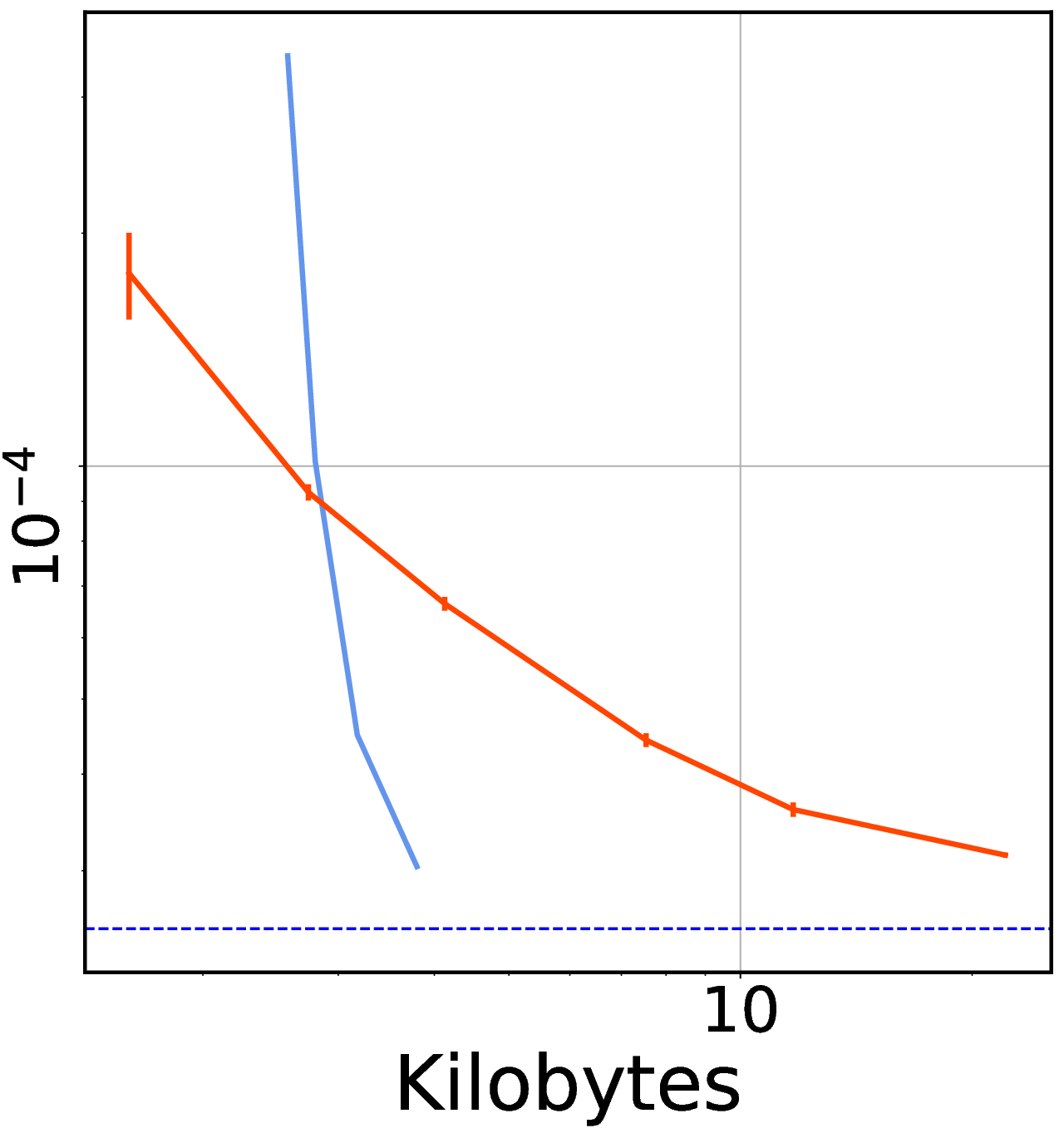} &
      \includegraphics[width=\fourPlotsSize]{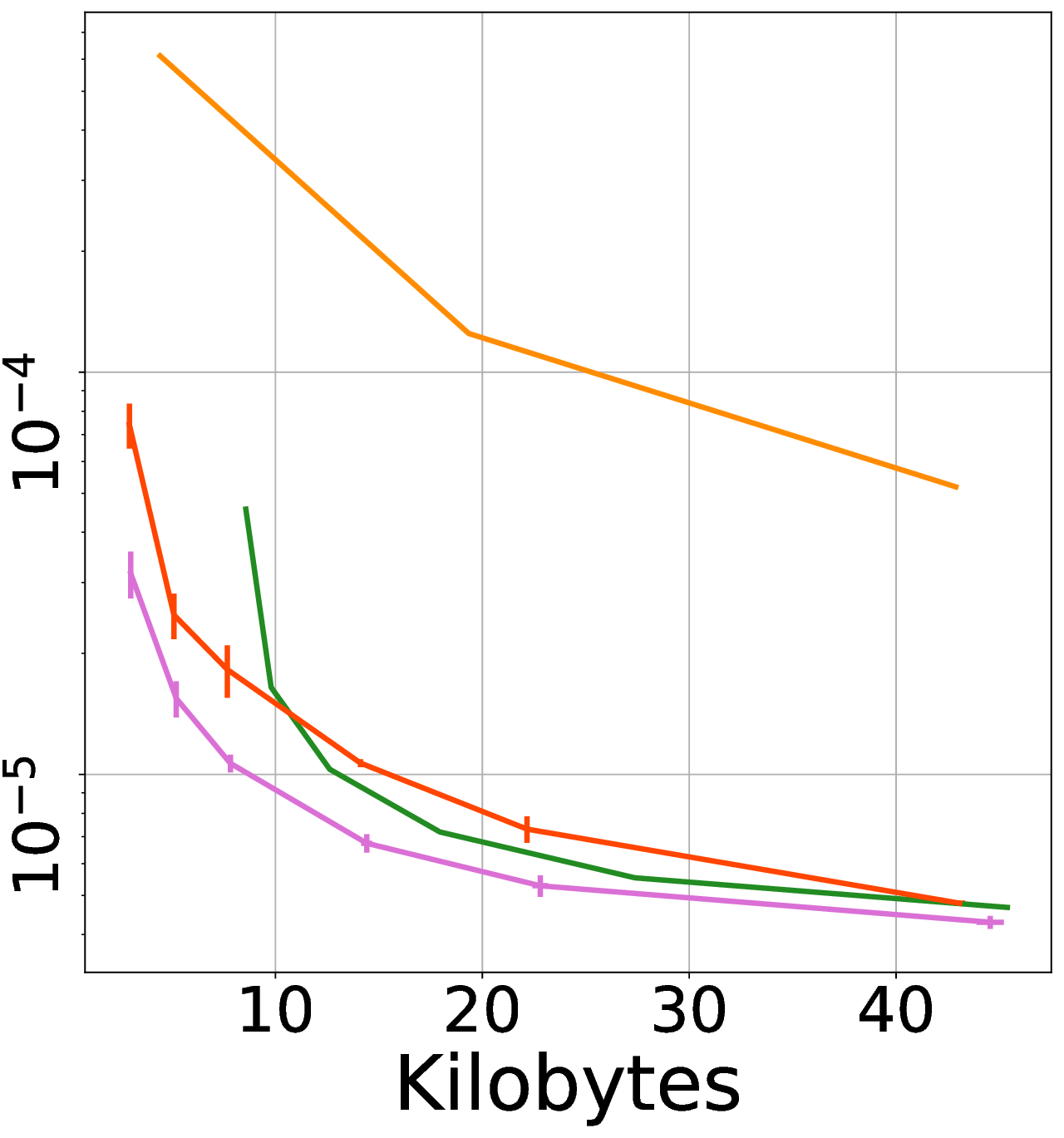} &
      \includegraphics[width=\fourPlotsSizeWithLabel]{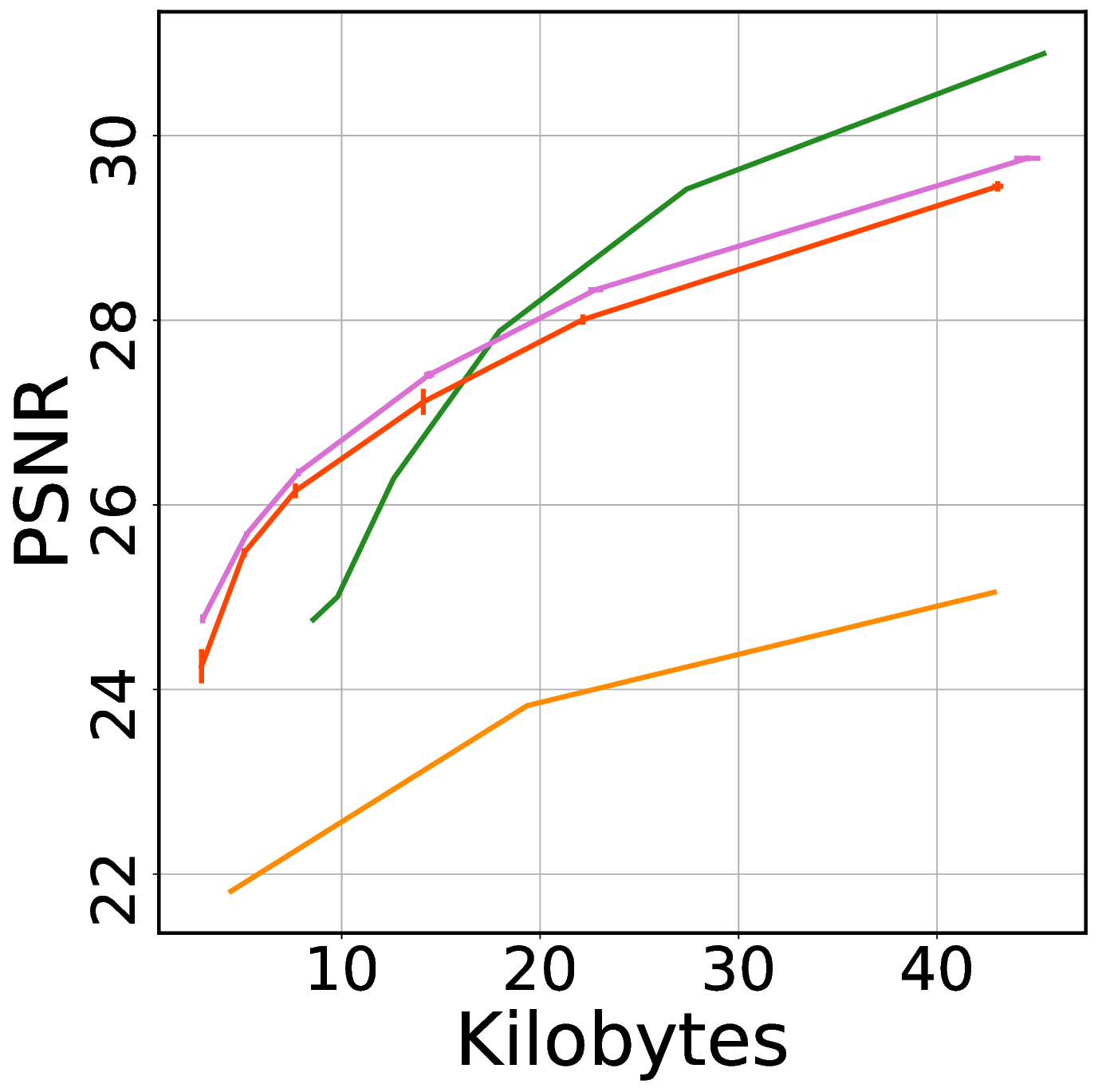} \\
      a) Stanford &  b) Garments & c) Geometry &  d) Color Attributes \\
\end{tabular}
\caption{Rate-distortion plot for \textbf{mesh compression} on the \stanford{}~\cite{turk1994zippered} (a) and the \garments{}~\cite{bhatnagar2019mgn} (b), and for \textbf{geometry and attribute compression} (c/d) on \attr{}~\cite{d20178i}. On mesh compression, we report the average \ac{cd} and kilobytes for \acp{nf} based on \acp{udf}/\acp{sdf} and Draco. We do not evaluate the performance of \acp{nf} using \acp{sdf} due to non-watertight shapes. In (c)/(d), we report the average \ac{cd}/PSNR and kilobytes for \acp{nf}-based on \acp{udf}/\acp{sdf}, GPCC and VPCC. There is no theo. min. in (c) as we use the \acp{pc}.}\label{fig:geometry_mesh_stanford_garments}
\end{figure*}

\textbf{Meshes.} \autoref{fig:geometry_mesh_stanford_garments} (a) \& (b) contains quantitative mesh compression results on the \stanford{} and the \garments{}. 
Moreover, we refer to the supplement for qualitative results of reconstructed meshes.
We only evaluate \acp{udf} since both datasets contain non-watertight shapes.
We find that \ac{nf}-based compression outperforms Draco on more complex meshes (\stanford{}), while Draco can outperform \acp{udf} on simpler meshes (\garments{}).
This is reasonable since the meshes in the \garments{} contain large planar regions which benefits the edge-breaker algorithm~\cite{rossignac1999edgebreaker} used by Draco as it is easier to represent such regions with only a few triangles. 

\begin{table}
  \caption{We report the average encoding (top) and decoding (bottom) runtime in [s] on the Stanford shape repository of Draco, \ac{pcgcv2}, GPCC, VPCC, VQAD and our approach based on \acp{udf}. Draco, GPCC and VPCC are evaluated on a CPU, while \ac{nf}-based compression, VQAD and PCGCv2 are evaluated on a single V100.}\label{tab:enc_dec_runtime}
  \label{sample-table}
  \centering
  \begin{tabular}{cccccc}
    \toprule
    Draco & PCGCv2 & GPCC & VPCC & VQAD & OURS\\
    \midrule
    $0.06$ & $0.50$ & $0.26$  &  $46.1$  & $172$ & $500$ \\
    $0.04$ & $0.96$ & $0.04$ &  $2.3$  & $3.3$ & $1.6$ \\
    \bottomrule
  \end{tabular}
\end{table}

\begin{figure*}
\centering
\setlength{\tabcolsep}{1pt}
\begin{tabular}{cccc}
      \includegraphics[width=\fourPlotsSizeWithLabel]{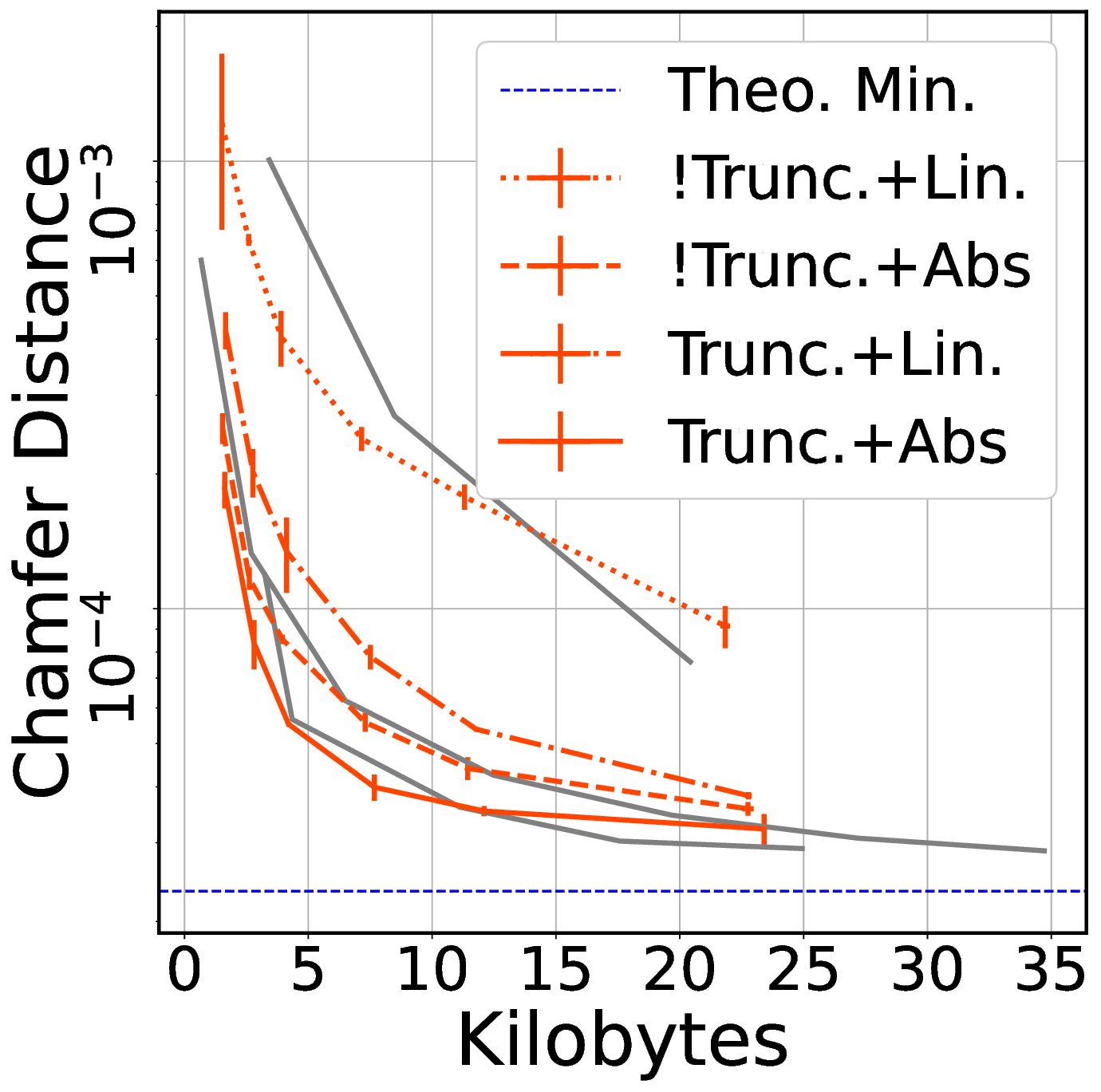} &
      \includegraphics[width=\fourPlotsSize]{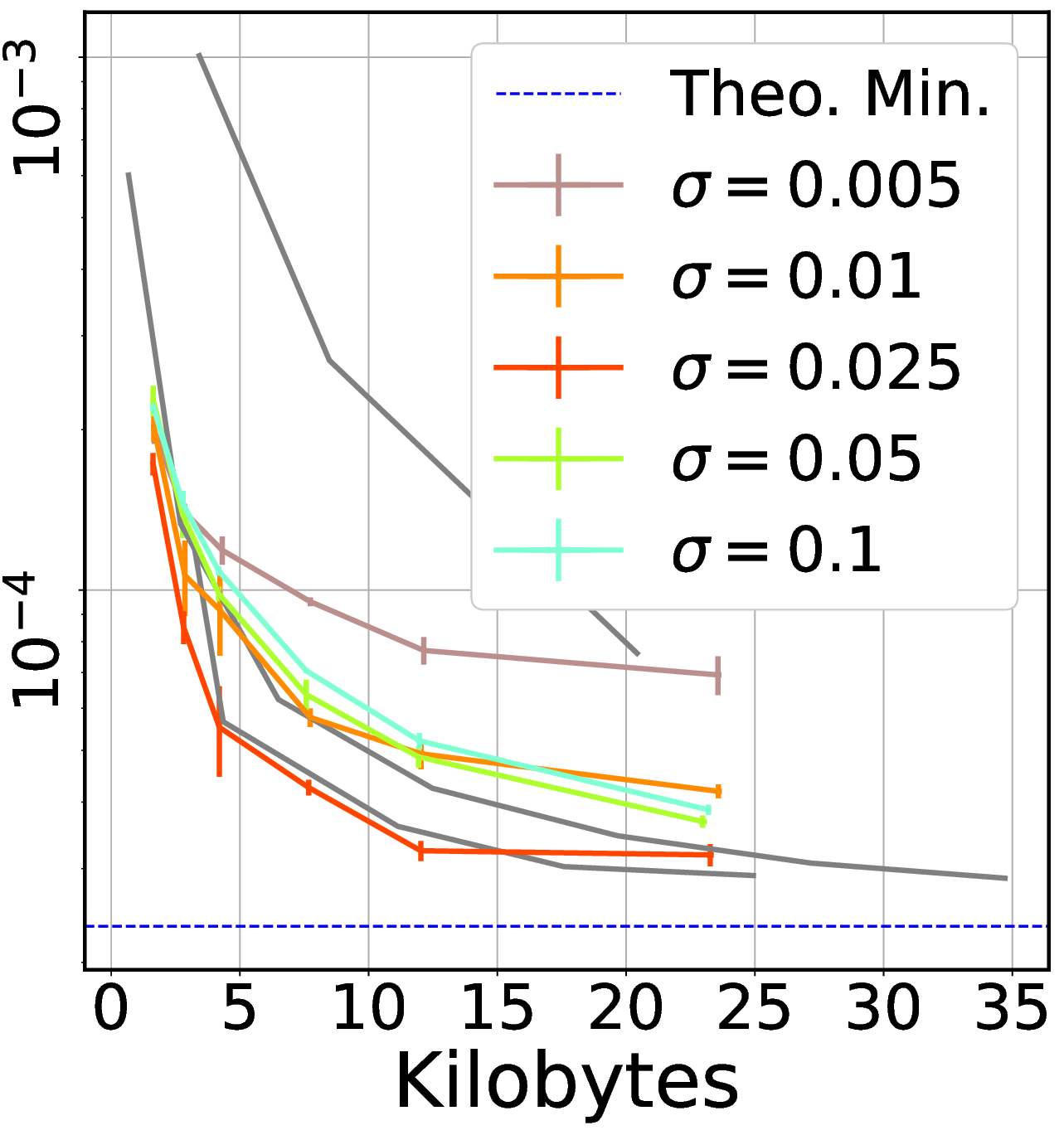} &
      \includegraphics[width=\fourPlotsSize]{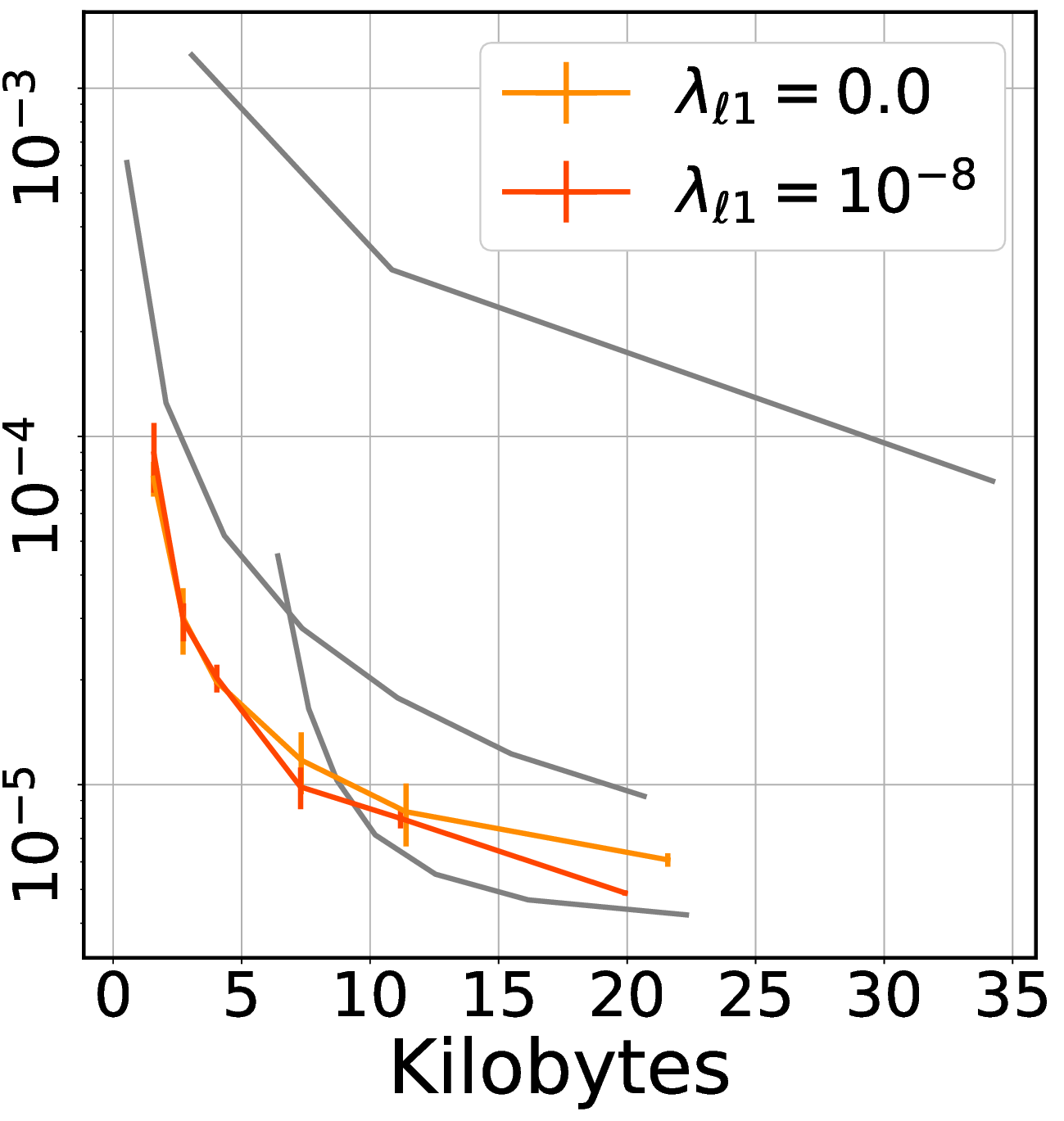} &
      \includegraphics[width=\fourPlotsSizeWithLabel]{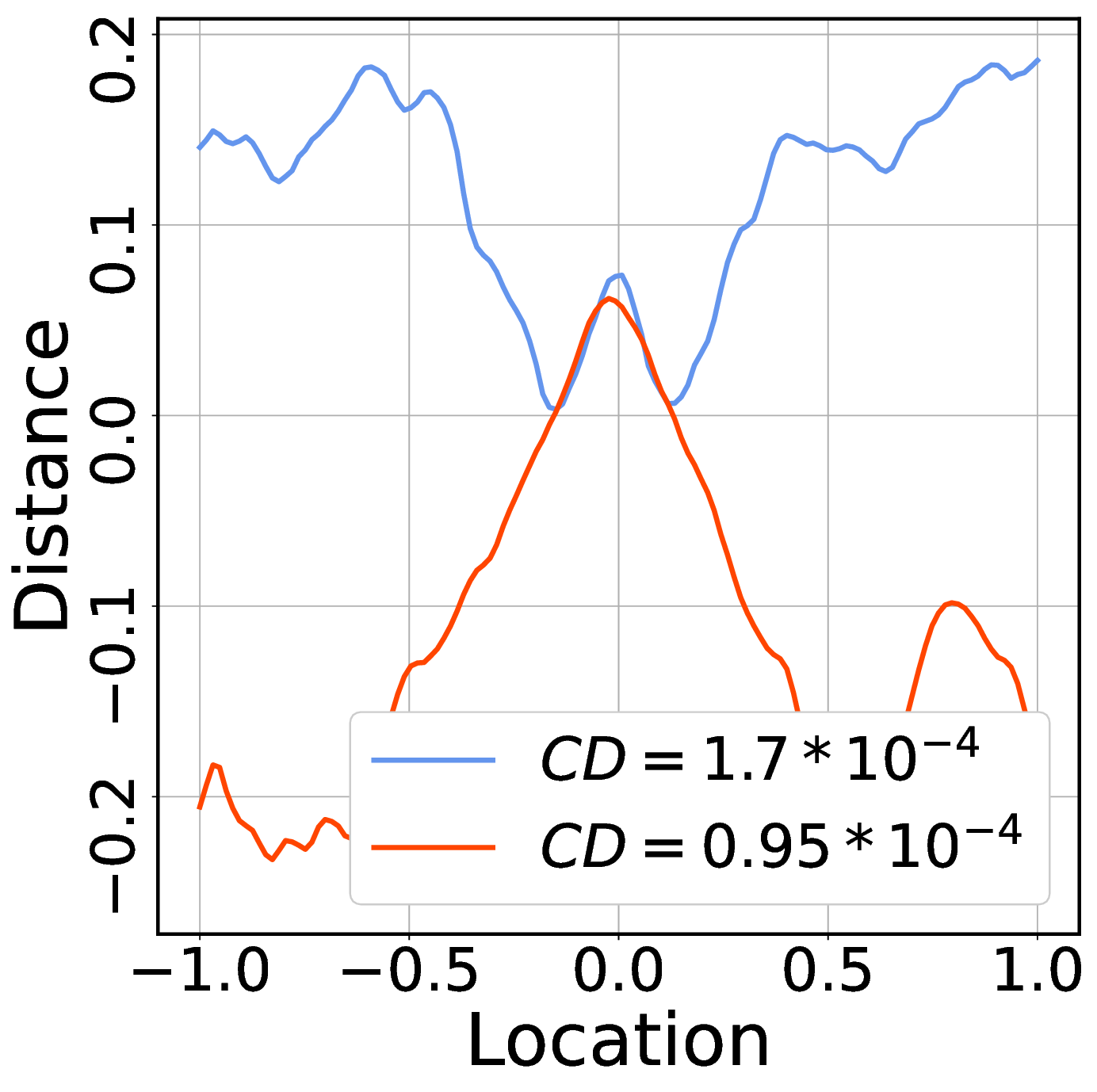} \\
      \small a) Truncation / Activation &  b) $\sigma$ &  c) $\ell 1$ & d) \ac{udf} w/o abs \\
\end{tabular}
\caption{
Impact of using truncated (Trunc.) \acp{df} and the abs activation function (a), regularization effect of the standard deviation $\sigma$ added to points during training (b) on the \stanford{}~\cite{turk1994zippered}. (c) depicts the impact of the $\ell 1$ penalty on \attr{}. We plot Draco, VPCC and GPCC in the background for reference (grey). We observe that \acp{udf} demonstrate the strongest performance when combining truncated unsigned distance fields with the abs activation function (a), when adding noise of standard deviation $\sigma=0.025$ to the raw \ac{pc} (b) and when applying an $\ell 1$-penalty to its parameters. We further visualize a 1D-cut through the center of the shape \textit{Armadillo} of the dataset \stanford{} for two independently trained \acp{udf} with a hidden dimension of 16 (d). We observe that the \ac{udf} which converges to a \ac{sdf} (red) prior to the abs activation function yields lower \ac{cd}.}\label{fig:geometry_pc_ablation_architecure_regularization}
\end{figure*}

\textbf{Architecture Choices.} We investigate the impact of using truncated \acp{df} and applying the abs activation function to the output of \acp{udf}. 
\autoref{fig:geometry_pc_ablation_architecure_regularization} (a) depicts the result. 
We observe that both, truncation and abs activation function, are essential for \acp{udf} with strong compression performance. Note that Chibane \etal{}~\cite{chibane2020neural} used ReLU as the final activation function. 
We refer to the supplement for a demonstration that ReLU performs similar to a linear activation function. 

\textbf{Regularization.} We demonstrate the impact of regularization in \autoref{fig:geometry_pc_ablation_architecure_regularization} (b) \& (c).
Adding Gaussian noise $\sigma \sim \mathcal{N}(0; 0.025)$ and applying an $\ell 1$-penalty to the parameters $\theta_G$ increasingly improves performance for larger \acp{nf}. 
This is expected as larger \acp{nn} require more regularization to prevent overfitting which is a problem for \ac{nf}-based compression of 3D geometries - in contrast to other data modalities.

\textbf{UDF Parameter Initialization.} Interestingly, we observe that the \ac{cd} of \ac{nf}-based compression using \acp{udf} has a large variance.
In fact, the primary source of this randomness is the parameter initialization which we find by optionally fixing the random seed of the dataset and parameter initialization.
For this result we refer the reader to the supplement.
Moreover, qualitatively we find in \autoref{fig:geometry_pc_ablation_architecure_regularization} (d) that \acp{udf} which converge to a \acp{sdf} prior to the abs activation function yield lower \ac{cd}.


\textbf{Runtime.} We compare the encoding and decoding runtime of \ac{nf}-based compression with the baselines (see \autoref{tab:enc_dec_runtime}). 
The encoding runtime of our approach is slower compared to the baselines since it needs to fit a \ac{nn} to each instance. 
Notably, this can be improved using meta-learned initializations~\cite{strumpler2022implicit}
Interestingly, compared to \ac{vpcc}, which is the only baseline that is competitive in terms of compression performance on weaker compression ratios, \ac{nf}-based compression is competitive if decoding is performed on a GPU.
This renders \ac{nf}-based compression practical when a 3D shape needs to be decoded many more times than encoded.

    

\subsection{Attribute Compression}\label{ssec:attribute_compression}

\begin{figure*}
\centering

    \renewcommand{\arraystretch}{1}
    \newcommand{\imgwidth}{0.23}
    \newcommand{\spyloc}{(-0.9, 0)}
    \newcommand{\spypos}{(0., 1.2)}
    \newcommand{\spysize}{1.0cm}
    \begin{tabular}{ccccc}    
    \begin{tikzpicture}[spy using outlines={circle,purple,magnification=2,size=\spysize, connect spies}]
			\node {\includegraphics[width=\attributesPCQualitativeFigWidth]{"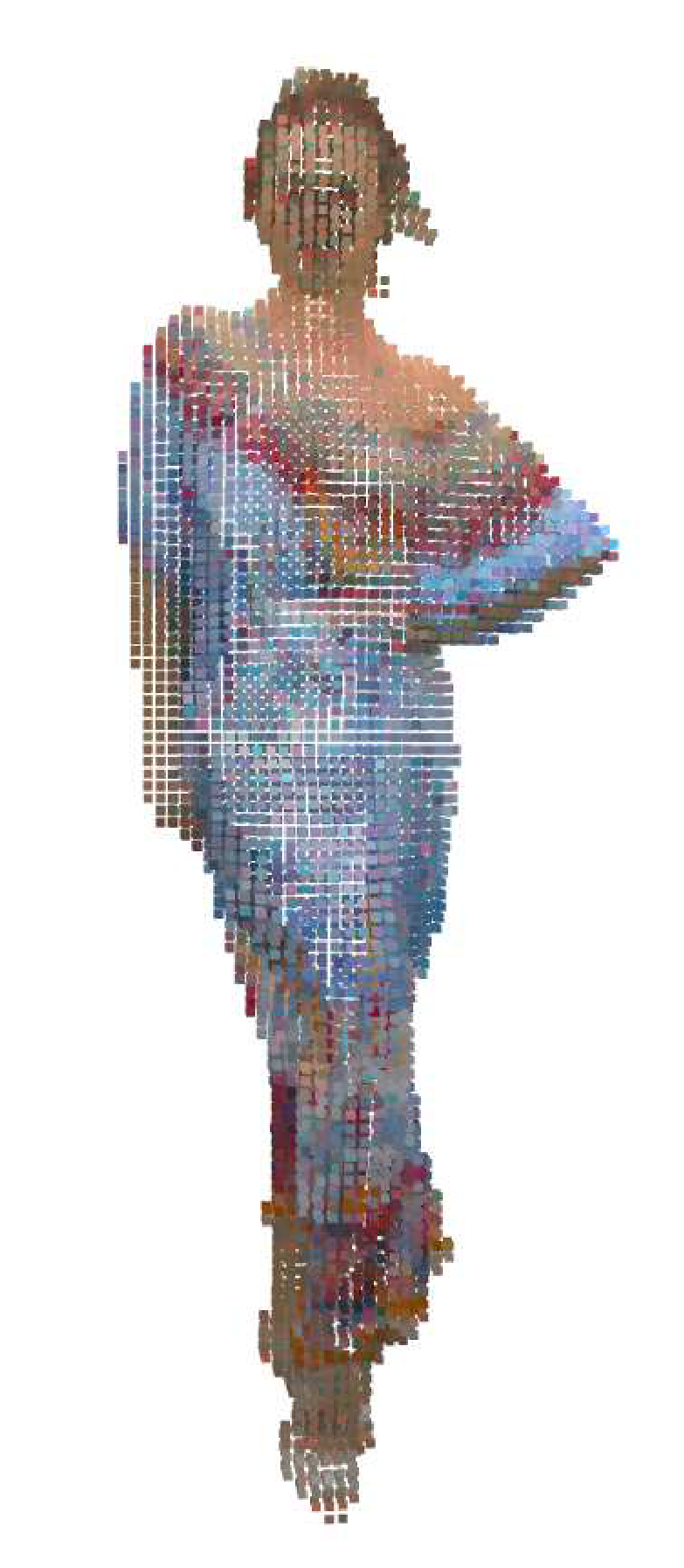"}};
			\spy on \spypos in node [] at \spyloc;
    \end{tikzpicture} & 
    \begin{tikzpicture}[spy using outlines={circle,purple,magnification=2,size=\spysize, connect spies}]
			\node {\includegraphics[width=\attributesPCQualitativeFigWidth]{"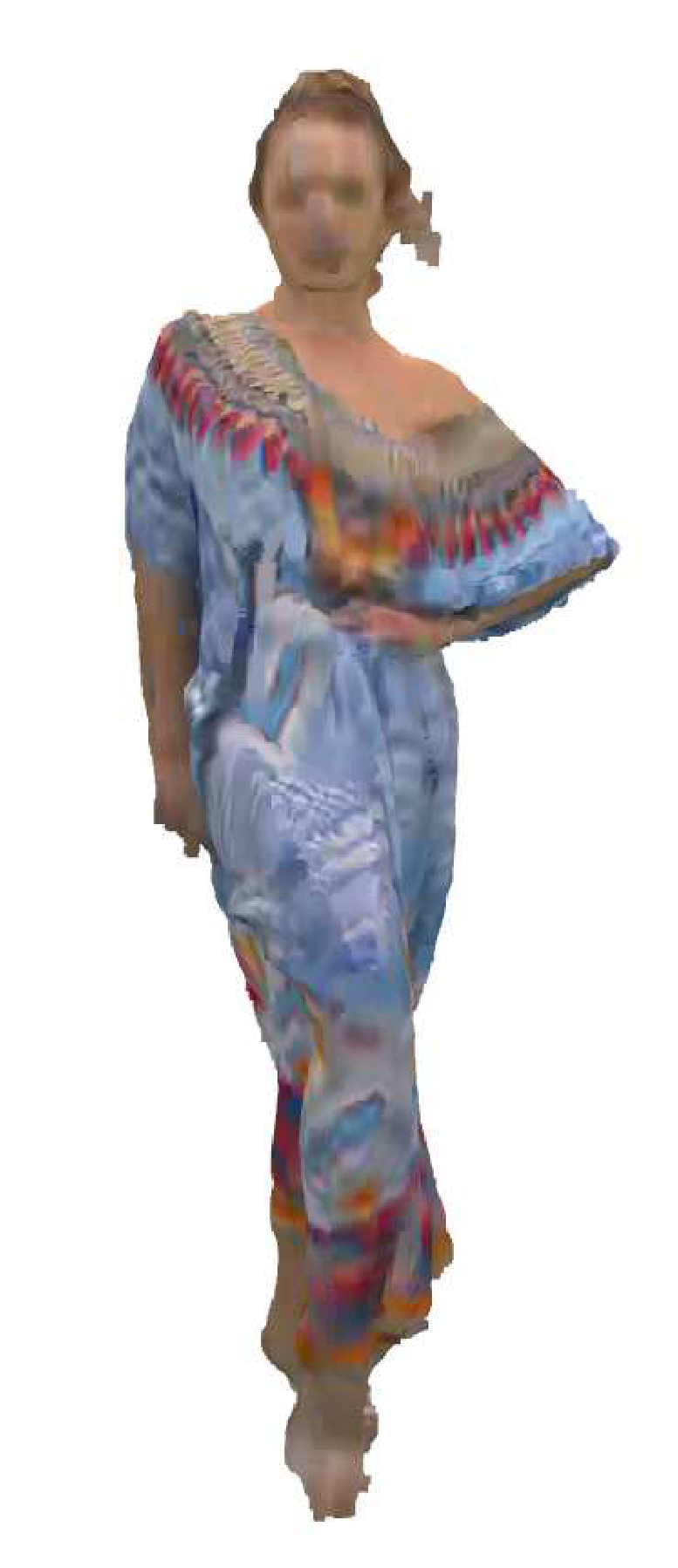"}};
			\spy on \spypos in node [] at \spyloc;
    \end{tikzpicture} & 
    \begin{tikzpicture}[spy using outlines={circle,purple,magnification=2,size=\spysize, connect spies}]
			\node {\includegraphics[width=\attributesPCQualitativeFigWidth]{"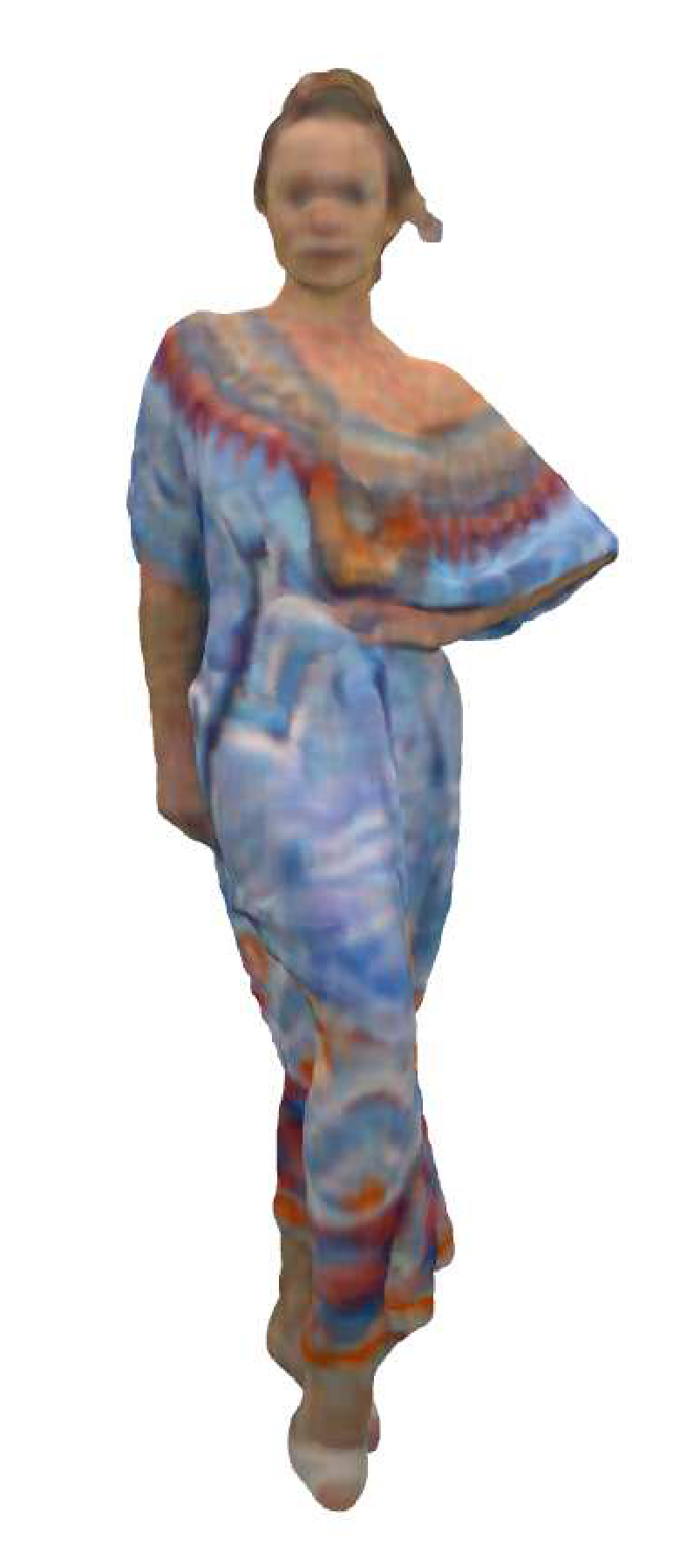"}};
			\spy on \spypos in node [] at \spyloc;
    \end{tikzpicture} & 
    \begin{tikzpicture}[spy using outlines={circle,purple,magnification=2,size=\spysize, connect spies}]
			\node {\includegraphics[width=\attributesPCQualitativeFigWidth]{"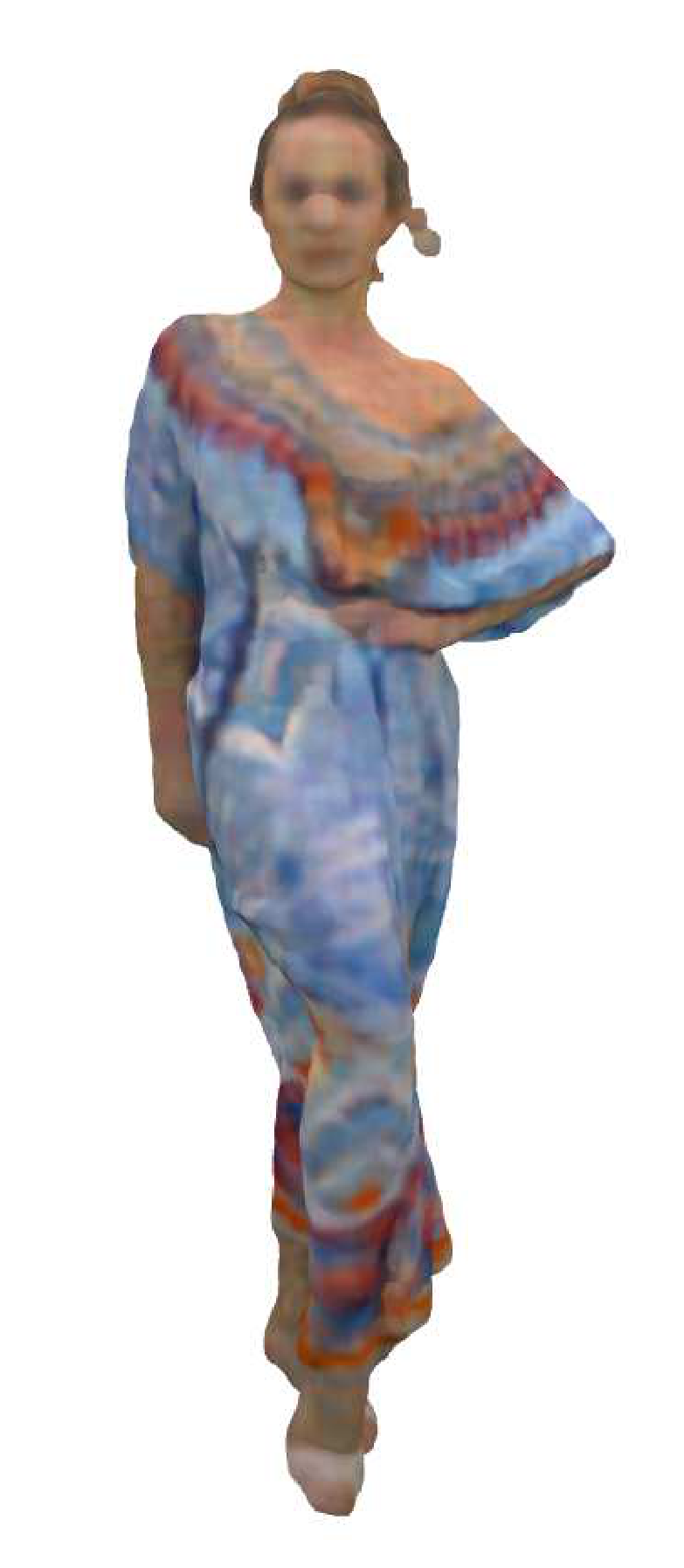"}};
			\spy on \spypos in node [] at \spyloc;
    \end{tikzpicture} & 
    \begin{tikzpicture}[spy using outlines={circle,purple,magnification=2,size=\spysize, connect spies}]
			\node {\includegraphics[width=\attributesPCQualitativeFigWidth]{"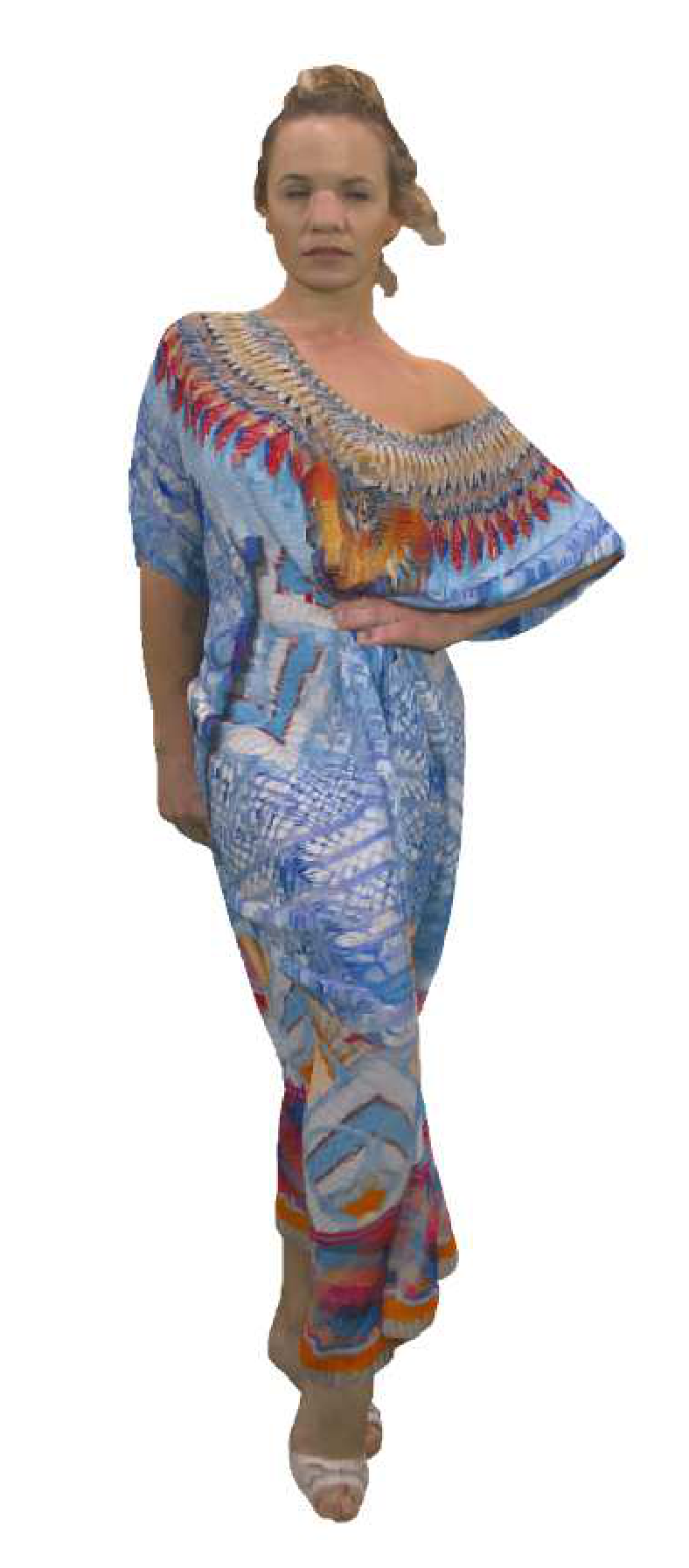"}};
			\spy on \spypos in node [] at \spyloc;
    \end{tikzpicture} 
    \\ \small a) GPCC (19.3 KB) & \small b) VPCC (18.0 KB) & \small c) \ac{udf} (15.7 KB) & \small d) \ac{sdf} (15.7 KB) & \small e) GT (20.2 MB)\\
    \end{tabular}
\caption{"Longdress" of \attr{} for \ac{gpcc} (a), \ac{vpcc} (b), \acp{udf}/\acp{sdf} (c/d) and the target (e).}\label{fig:geometry_attributes_qualitative}
\end{figure*}

Furthermore, we evaluate \ac{nf}-based compression on \acp{pc} containing color attributes on \attr{} and compare it with \ac{gpcc} and \ac{vpcc}.
\autoref{fig:geometry_mesh_stanford_garments} (c) \& (d) show quantitative and \autoref{fig:geometry_attributes_qualitative} qualitative results.
\ac{nf}-based compression using \ac{sdf} outperforms both baselines on geometry compression. 
\acp{udf} outperform \ac{vpcc} for strong compression ratios.
On attribute compression \acp{sdf} and \acp{udf} outperform \ac{gpcc} by a large margin, but \ac{vpcc} only for stronger compression ratios.
We show in the supplement that jointly compressing geometry and attributes leads to worse performance.

\setlength{\tabcolsep}{12pt} 

\section{Discussion}\label{sec:conclusion}
We proposed an algorithm for compressing the geometry as well as attributes of 3D data using \acp{nf}.
We introduced two variants - one specialized to watertight shapes based on \acp{sdf} and another more general approach based on \acp{udf} (see \autoref{sec:method}).
For watertight shapes \acp{sdf} demonstrate strong geometry compression performance across all compression ratios (see \autoref{ssec:geometry_compression}). 
\acp{udf} perform well on geometry compression - in particular for stronger compression ratios.
However, \ac{vpcc} can outperform \acp{udf} on weaker compression ratios. 
Notably, the learned neural baseline \acp{pcgcv2} shows strong performance on \attr{}, but suffers from large performance drops on \stanford{} and \garments{}.
This highlights another strength of \ac{nf}-based neural compression - it does not exhibit the sensitivity to domain shifts of other neural compression algorithms. 
On attribute compression (see \autoref{ssec:attribute_compression}) \acp{sdf} as well as \acp{udf} outperform both baselines for stronger compression ratios, while \ac{vpcc} performs better when less compression is required.

Interestingly, we observed in \autoref{ssec:geometry_compression} that the decoding runtime of \ac{nf}-based compression is competitive on a GPU.
This is in line with recent findings on \ac{nf}-based video compression~\cite{lee2022ffnerv}.
However, the encoding runtime remains one order of magnitude larger than the next slowest method (\ac{vpcc}). 
This gap can potentially be closed when using meta-learned initializations which have been found to improve \ac{nf}-based image compression~\cite{strumpler2022implicit,dupont2022coin++} and speed up convergence by a factor of 10~\cite{strumpler2022implicit}.

Furthermore, we found that the performance of \acp{udf} strongly depends on the parameter initialization (see \autoref{ssec:geometry_compression}). 
When using the abs activation function, \acp{udf} are flexible regarding the values they predict prior to it.
On watertight shapes,  \acp{udf} can converge to a function predicting the same sign inside and outside or one that predicts different signs. 
%
%
In presence of a sign flip, \acp{udf} perform better. 
Thus, a promising direction for improving the more general compression method based on \acp{udf} are novel initialization methods beyond the one provided in Sitzmann \etal{}~\cite{sitzmann2020implicit}. 
Notably, one line of work aims at learning \acp{sdf} form raw point clouds by initializing \acp{nf} such that they approximately resemble an \ac{sdf} of an r-ball after initialization~\cite{atzmon2020sal, atzmon2020sald}.
However, none of these methods work when using positional encodings which are necessary for strong performance.


{
    \small
    \bibliographystyle{ieeenat_fullname}
    \bibliography{main}
}


\end{document}